\documentclass[10pt,journal]{IEEEtran}

\usepackage{cite}
\usepackage{graphics} 
\usepackage{epsfig} 
\usepackage{epstopdf}
\usepackage{mathrsfs}
\usepackage{times} 
\usepackage{amsmath} 
\usepackage{amssymb}  
\usepackage{epstopdf}
\usepackage{subfigure}
\usepackage{epic}
\usepackage{bm}
\usepackage{tikz}
\usetikzlibrary{arrows,shapes,snakes,automata,backgrounds,petri}
\usepackage{algorithm}
\usepackage{algpseudocode}
\usepackage{url}
\usepackage{booktabs, caption, makecell}

\usepackage{threeparttable}

\usepackage{graphicx} 
\usepackage{color} 

\floatname{algorithm}{Algorithm}

\newtheorem{definition}{Definition}
\newtheorem{theorem}{Theorem}
\newtheorem{lemma}{Lemma}

\newtheorem{remark}{Remark}
\newtheorem{assumption}{Assumption}
\newtheorem{proposition}{Proposition}
\newtheorem{example}{Example}

\begin{document}
%
\title{Distributed motion coordination for multi-robot systems under LTL specifications}
%
%
%

\author{Pian Yu and Dimos V. Dimarogonas
\thanks{This work was supported in part by the Swedish Research Council (VR), the Swedish Foundation for Strategic Research (SSF) and the Knut and Alice Wallenberg Foundation (KAW).}
\thanks{The authors are with School of Electrical Engineering and Computer Science, KTH Royal Institute of Technology, 10044 Stockholm, Sweden.
        {\tt\small piany@kth.se, dimos@kth.se}}
}

\maketitle

\begin{abstract}
This paper investigates the online motion coordination problem for a group of mobile robots moving in a shared workspace, each of which is assigned a linear temporal logic specification. Based on the realistic assumptions that each robot is subject to both state and input constraints and can have only local view and local information, a fully distributed multi-robot motion coordination strategy is proposed. For each robot, the motion coordination strategy consists of three layers. An offline layer pre-computes the braking area for each region in the workspace, the controlled transition system, and a so-called potential function. An initialization layer outputs an initially safely satisfying trajectory. An online coordination layer resolves conflicts when one occurs. The online coordination layer is further decomposed into three steps. Firstly, a conflict detection algorithm is implemented, which detects conflicts with neighboring robots. Whenever conflicts are detected, a rule is designed to assign dynamically a planning order to each pair of neighboring robots. Finally, a sampling-based algorithm is designed to generate local collision-free trajectories for the robot which at the same time guarantees the feasibility of the specification. Safety is proven to be guaranteed for all robots at any time. The effectiveness and the computational tractability of the resulting solution is verified numerically by two case studies.
\end{abstract}


\IEEEpeerreviewmaketitle

\section{Introduction}

\IEEEPARstart{O}{ne} challenge for multi-robot systems (MRSs) is the design of coordination strategies between robots that enable them to perform operations safely and efficiently in a shared workspace while achieving individual/group motion objectives \cite{yan2013survey}. This problem was originated from the 80s and has been extensively investigated since. In recent years, the attention that has been put on this problem has grown significantly due to the emergence of new applications, such as smart transportation and service robotics. The existing literature can be divided into two categories: \emph{path coordination} and \emph{motion coordination}. The former category plans and coordinates the entire paths of all the robots in advance (offline), while the latter category focuses on decentralized and online approaches that allow robots to resolve conflicts online when one occurs\footnote{In some literature these two terms are used interchangeably. In this paper, we try to distinguish between the two as explained above.} \cite{parker2009path}. This paper aims at developing a fully distributed strategy for multi-robot motion coordination (MRMC) with safety guarantees.

Depending on how the controller is synthesized for each robot, the literature concerning MRMC can be further classified into two types: the reactive approach and the planner-based approach. Typical methods that generate reactive controllers consist of the potential-field approach \cite{khatib1986real, panagou2014motion}, sliding mode control \cite{gracia2013reactive,farivarnejad2016decentralized} and control barrier functions~\cite{wang2017safety,glotfelter2017nonsmooth}. These reactive-style methods are fast and operate well in real-time. However, it is well-known that these methods are sensitive to deadlocks that are caused by local minima. Moreover, although these reactive-style methods work well in relatively unconstrained situations, guidance for setting control parameters is not analyzed formally when explicit constraints on the system states and/or inputs are presented \cite{parker2009path}. Apart from the above, other reactive methods include the generalized roundabout policy \cite{pallottino2007decentralized} and a family of biologically inspired methods \cite{bekey2005autonomous}.

An early example of the planner-based method is the work of Azarm and Schmidt \cite{azarm1997conflict}, where a framework for online coordination of multiple mobile robots was proposed. In this framework, MRMC was solved as a sequential trajectory planning problem, where priorities are assigned to robots when conflicts are detected, and then a motion planning algorithm is implemented to generate conflict-free paths. Based on this framework, various applications and different motion planning algorithms are investigated. Guo and Parker \cite{guo2002distributed} proposed a MRMC strategy based on the D$^*$ algorithm. In this work, each robot has an independent goal position to reach and know all path information. In \cite{sheng2006distributed}, a distributed bidding algorithm was designed to coordinate the movement of multiple robots, which focuses on area exploration. In the work of Liu \cite{liu2017distributed}, conflict resolution at intersections was considered for connected autonomous vehicles, where each vehicle is required to move along a pre-planned path. A literature review on MRMC can be found in \cite{yan2013survey}.

No matter which type of controllers is implemented, safety has always been a crucial issue for MRMC. In \cite{wang2017safety}, MRSs with double-integrator dynamics were studied, control barrier functions were proposed to solve the motion coordination problem and safety guarantees were established. However, the velocity constraints are not dealt with. In \cite{liu2017distributed}, safety was stated by assuming that the deceleration that each robot can take is unbounded, yet this assumption may not be realistic for practical applications. In addition, most of the above mentioned literature concerning MRMC considers relatively simple tasks for each robot (\textit{e.g.,} an arrival task from initial state to goal state). However, as robots become more capable, a recent trend in the area of robot motion planning is to assign robots more complex, high-level tasks, such as temporal logic specifications.

In the last few years, multi-robot control under linear temporal logic (LTL) specifications has been investigated in \cite{ulusoy2013optimality,shoukry2017linear,sahin2019multirobot,kantaros2020stylus,alonso2018reactive,guo2015multi}. Most of them consider that the MRS is subject to a global LTL specification, and then an offline motion planning problem is solved in a centralized manner \cite{ulusoy2013optimality,shoukry2017linear,sahin2019multirobot,kantaros2020stylus}. In \cite{alonso2018reactive}, unknown moving obstacles were taken into account. Therefore, online coordination with the moving obstacles was further required. In this work, safety is shown under the assumptions that each robot is moving at a constant velocity and that the local motion planning is feasible. In \cite{guo2015multi}, multi-robot plan reconfiguration under local LTL specifications was investigated. A potential-field-based navigation controller was implemented for each robot to guarantee safety. However, the approach is not applicable when state and input constraints are considered.

Motivated by the above observations, this paper investigates the MRMC problem for a group of mobile robots moving in a shared workspace, each of which is assigned a LTL specification. Robots are assumed to have limited sensing capabilities and constraints in both state and input are considered. To cope with these setups, a fully distributed MRMC strategy is proposed. The contributions of this paper are summarized as follows.
\begin{itemize}
  \item [i)] A framework for distributed MRMC under LTL specifications is proposed for each robot, which consists of three layers: an offline pre-computation layer, an initialization layer, and an online coordination layer. The online coordination layer is further decomposed into three steps. Firstly, conflicts are detected within the sensing area of each robot. Once conflicts are detected, a rule is applied to assign dynamically a planning order to each pair of neighboring robots. Finally, a sampling-based algorithm is implemented for each robot that generates a local collision-free trajectory which at the same time satisfies the LTL specification.
  \item [ii)] Safety is established under all circumstances by combining the planner-based controller with an emergency braking controller.
  \item [iii)] As the motion coordination strategy is designed to be fully distributed and each robot considers only local information of neighboring robots, it is totally scalable in the sense that the computational complexity of the strategy does not increase with the number of robots in the workspace.
\end{itemize}
A comparison between this work and the related literature \cite{ulusoy2013optimality,shoukry2017linear,sahin2019multirobot,kantaros2020stylus,guo2015multi,alonso2018reactive} is summarized in Table I.


\begin{table}[]
\centering
 \begin{threeparttable}[b]
\caption{Comparison of this work to related literature.}
\begin{tabular}{c||c|c|c|c}
\hline
\bf{Literature} & \bf{Task}\tnote{1} & \begin{tabular}[c]{@{}c@{}}\bf{Plan} \\ \bf{synthesis}\end{tabular} & \begin{tabular}[c]{@{}c@{}}\bf{Online} \\ \bf{coordination}\end{tabular} & \bf{Safety}\tnote{2}           \\ \hline
{\cite{ulusoy2013optimality,shoukry2017linear,sahin2019multirobot,kantaros2020stylus}}     & global      & centralized      & no                  & \begin{tabular}[c]{@{}c@{}}offline  safety\end{tabular}           \\ \hline
{\cite{alonso2018reactive}}     & global        & centralized      & yes                 & \begin{tabular}[c]{@{}c@{}}additional \\ assumptions\end{tabular}    \\ \hline
{\cite{guo2015multi}}     &local        & distributed      & yes                 & \begin{tabular}[c]{@{}c@{}}no state and \\ input constraints \\ are considered\end{tabular}    \\ \hline
{This work}     & local        & distributed      & yes                 & \begin{tabular}[c]{@{}c@{}}no additional \\ assumptions, \\ state and input \\ constraints \\ are considered\end{tabular}  \\ \hline
\end{tabular}
\begin{tablenotes}
     \item[1] Global means that a team LTL specification is assigned to all robots, local means that an individual LTL specification is assigned to each robot.
      \item[2] Safety means robot-robot and robot-obstacle collision avoidance.
   \end{tablenotes}
  \end{threeparttable}
\end{table}

A preliminary version of this work was accepted by the
2020 American Control Conference \cite{yu2020fully}. Here, we expand this preliminary version in two main directions. Firstly, the framework is generalized
to LTL specifications. In the conference version, only reach-avoid type of tasks are considered, and the replanning problem can be formulated as an optimization problem. However, as the verification of an LTL formula is in general difficult to be conducted online, a local trajectory generation algorithm is designed in this work. Secondly, safety guarantees are established without the assumption that each robot can take unbounded input in case of emergency. This is done by considering a braking distance at both the conflict detection and coordination steps.

The remainder of the paper is organized as follows. In Section II, notation and preliminaries on transition systems, LTL and product automaton are introduced. Section III formalizes the (online) motion coordination problem. Section IV presents the proposed solution in detail, which is verified by two case studies in Section V. A summary of this work is given in Section VI.

\section{Preliminaries}

\subsection{Notation}
Let $\mathbb{R}:=(-\infty, \infty)$, $\mathbb{R}_{\ge 0}:=[0, \infty)$, $\mathbb{R}_{> 0}:=(0, \infty)$, and $\mathbb{N}:=\{0,1,2,\ldots\}$. Denote $\mathbb{R}^n$ as the $n$-dimensional real vector space, $\mathbb{R}^{n\times m}$ as the $n\times m$ real matrix space. Throughout this paper, vectors are denoted in italics, $x\in \mathbb{R}^n$, and boldface $\bm{x}$ is used for continuous-time signals or a sequence of states. Given a continuous-time signal $\bm{x}$, $x(t)$ denotes the value of $\bm{x}$ at time $t$.
$0_n$ denotes a $n$-dimensional column vector
with all elements equal to 0. $[a, b]$ and $[a, b)$ denote closed and right half-open intervals with end points $a$ and $b$. For $x_1\in\mathbb{R}^{n_1}, \ldots, x_m\in\mathbb{R}^{n_m}$, the notation $(x_1, x_2, \ldots, x_m)\in \mathbb{R}^{n_1+n_2+\cdots +n_m}$ stands for $[x_1^T, x_2^T, \ldots, x_m^T]^T$. Let $\left|\lambda\right|$ be the absolute value of a real number $\lambda$, $\|x\|$ and $\|A\|$ be the Euclidean norm of vector $x$ and matrix $A$, respectively. Given a set $\Omega$, $2^{\Omega}$ denotes its powerset and $|\Omega|$ denotes its cardinality. Given two sets $\Omega_1, \Omega_2$, the set $\mathcal{F}(\Omega_1, \Omega_2)$ denotes the set of all functions from $\Omega_1$ to $\Omega_2$. The operators $\cup$ and $\cap$ represent set union and set intersection, respectively. In addition, we use $\wedge$ to denote the logical operator AND and $\vee$ to denote the logical operator OR. The set difference $A\setminus B$ is defined by $A\setminus B:=\{x: x\in A \;\wedge\; x\notin B\}$.

Given a vector $x\in \mathbb{R}^{n}$, define the projection operator $\texttt{proj}_{m}(x): \mathbb{R}^{n} \to \mathbb{R}^{m}$ as a mapping from $x$ to its first $m, m\le n$ components. Given a signal $\bm{x}: [t_1, t_2]\to \mathbb{R}^{n}$, define $\texttt{proj}_{m}(\bm{x}):=\{\bm{x}': [t_1, t_2]\to \mathbb{R}^{m}| \bm{x}'(t)= \texttt{proj}_{m}(\bm{x}(t)), t\in [t_1, t_2]\}$. In addition, we use $\texttt{dom}(\bm{x})$ to represent the domain of a signal $\bm{x}$. Given two signals $\bm{x}_1: [t_1, t_2]\to \mathbb{R}^{n}$ and $\bm{x}_2: [t_2, t_3]\to \mathbb{R}^{n}$, denote by $\bm{x}_1\uplus \bm{x}_2:=\{\bm{x}: [t_1, t_3]\to \mathbb{R}^{n}| \bm{x}(t)= \bm{x}_i(t), t\in \texttt{dom}(\bm{x}_i), i={1, 2}\}$. Given a point $c\in \mathbb{R}^n$, a set $A\subseteq \mathbb{R}^{n}$ and a constant $r\ge 0$, $\texttt{dist}(c, A):=\inf_{y\in A}\{\|c-y\|\}$ represents the point-to-set distance and $\mathcal{B}(c, r)$ represents a ball area centered at point $c$ and with radius $r$. Denote by $\mathcal{B}(A, r):=\cup_{c\in A}\mathcal{B}(c, r)$. Given two sets $A, B$, define $\texttt{dist}(A, B):=\inf_{x\in A, y\in B}\{\|x-y\|\}$.

\subsection{Graph Theory}

Let $\mathcal{G}=\{\mathcal {V}, \mathcal {E}\}$ be a graph with the set of nodes $\mathcal {V}=\{1, 2,\dots, N\}$, and $\mathcal{E}\subseteq \{(i,j): i,j\in \mathcal {V},j \ne i\}$ the set of edges. If $(i,j)\in \mathcal {E}$, then node $j$ is called a neighbor of node $i$ and node $j$ can receive information from node $i$. The neighboring set of node $i$ is denoted by $\mathcal{N}_i = \{j \in \mathcal {V} | (i, j)\in \mathcal {E}\}$. Define $\mathcal{N}_i^+=\mathcal{N}_i\cup \{i\}$. The graph $\mathcal{G}$ is called undirected if $j\in \mathcal{N}_i \Rightarrow i\in \mathcal{N}_j, \forall j \ne i$. Given an edge $e_k:=(i, j)\in \mathcal {E}$, $i$ is called the head of $e_k$ and $j$ is called the tail of $e_k$. An undirected graph is called connected if for every pair of nodes $(i, j)$, there exists a path which connects $i$ and $j$, where a path is an ordered list of edges such that the head of each edge is equal to the tail of the following edge.

\subsection{LTL and B\"{u}chi automaton}

Let $AP$ be a set of atomic propositions. LTL is based on atomic propositions (state labels $a\in AP$), Boolean connectors like negation $\neg$ and conjunction $\wedge$, and two temporal operators $\bigcirc$ (``next") and $\mathsf{U}$ (``until"), and is formed according to the following syntax \cite{Baier2008}:
\begin{equation}\label{LTL}
  \varphi::=\text{true}|a|\neg\varphi|\varphi_1\wedge\varphi_2|\bigcirc \varphi|\varphi_1 \mathsf{U}\varphi_2,
\end{equation}
where $\varphi, \varphi_1, \varphi_2$ are LTL formulas. The Boolean connector disjunction $\vee$, and temporal operators $\lozenge$ (``eventually") and $\square$ (``always") can be derived as $\varphi_1\vee\varphi_2:=\neg(\neg\varphi_1\wedge \neg \varphi_2)$, $\lozenge\varphi:=\text{true} \mathsf{U}\varphi$ and $\square\phi:=\neg \lozenge\neg\phi$. Formal definitions for the LTL semantics and model checking can be found in \cite{Baier2008}.

\begin{definition}[B\"{u}chi automaton \cite{buchi1990decision}]\label{Buchi}
A nondeterministic B\"{u}chi automaton (NBA) is a tuple $\mathsf{B} = (S, S_{0}, 2^{AP}, \delta, F)$, where
\begin{itemize}
  \item $S$ is a finite set of states,
  \item $S_{0}\subseteq S$ is the set of initial states,
  \item $2^{AP}$ is the input alphabet,
  \item $\delta: S\times 2^{AP} \to 2^{S}$ is the transition function, and
  \item $F\subseteq S$ is the set of accepting states.
\end{itemize}
\end{definition}
An infinite \emph{run} $\bm{s}$ of a NBA is an infinite sequence of states $\bm{s}=s_0s_1\ldots$ generated by an infinite sequence of input alphabets ${\bm{\sigma}}=\sigma_0\sigma_1\ldots\in (2^{AP})^\omega$, where $s_0\in S_{0}$ and $s_{k+1}\in \delta(s_k, \sigma_k), \forall k\ge 0$. An infinite run $\bm{s}$ is called \emph{accepting} if $\texttt{Inf}(s)\cap F\neq \emptyset$, where $\texttt{Inf}(s)$ is the set of states that appear in $\bm{s}$ infinitely often. Given a state $s\in S$, define
\begin{equation}\label{post}
  \texttt{Post}(s):=\{s'\in S: \exists \sigma\in 2^{AP}, s'\in \delta(s, \sigma)\}.
\end{equation}
Given an LTL formula $\varphi$ over $AP$, there is a union of infinite words that satisfy $\varphi$, that is,
\begin{equation*}
  \texttt{Words}(\varphi)=\{\sigma\in (2^{AP})^\omega|\sigma\models \varphi\},
\end{equation*}
where $\models\subseteq (2^{AP})^\omega \times \varphi$ is the satisfaction relation \cite{Baier2008}.

\begin{lemma}\cite{gastin2001}
Any LTL formula $\varphi$ over $AP$ can be algorithmically translated into a B\"{u}chi automaton $\mathsf{B}_{\varphi}$ over the input alphabet $2^{AP}$ such that $\mathsf{B}_{\varphi}$ accepts all and only those infinite runs over $AP$ that satisfy $\varphi$.
\end{lemma}

\subsection{Transition system as embedding of continuous-time systems}

Consider a continuous-time dynamical system
\begin{equation}\label{x0}
\left\{\begin{aligned}
\dot{x}&=f(x, u),\\
      y&=g(x),
\end{aligned}\right.
\end{equation}
where $x\in X \subseteq \mathbb{R}^n$ is the state, $u\in U \subseteq \mathbb{R}^m$ is the control, $f: \mathbb{R}^n\times \mathbb{R}^m \to \mathbb{R}^n$ describes the dynamics, $y\in Y \subseteq \mathbb{R}^l$ is the output, and $g: \mathbb{R}^n\to \mathbb{R}^l$ is the output function.

Let $\mathcal{U}$ be the set of all measurable functions that take their values in $U$ and are defined on $\mathbb{R}_{\ge 0}$.
A curve ${\bm \xi}: [0, \tau) \to \mathbb{R}^n$ is said to be a trajectory of (\ref{x0}) if there exists an input signal ${\bm u}\in \mathcal{U}$ satisfying $\dot{\xi}(t)=f(\xi(t), u(t))$
for almost all $t\in [0, \tau)$. A curve ${\bm \zeta}: [0, \tau) \to \mathbb{R}^l$ is said to be an output trajectory of (\ref{x0}) if $\zeta(t)=g(\xi(t))$
for almost all $t\in [0, \tau)$, where ${\bm \xi}$ is a trajectory of (\ref{x0}). We use $\xi(\xi_0, {\bm u}, t)$ and $\zeta(\zeta_0, {\bm u}, t)$ to denote the trajectory and output trajectory point reached at time $t$ under the input ${\bm u}\in \mathcal{U}$ from initial condition $\xi_0$ and $\zeta_0$, respectively. In addition, when ${\bm u}$ is a constant signal, \textit{i.e.,} $u(t)\equiv \hat u, \forall t\in \texttt{dom}({\bm u})$ for a $\hat u\in U$, then we define $\xi(\xi_0, \hat u, t):=\xi(\xi_0, {\bm u}, t)$ and  $\zeta(\zeta_0, \hat u, t):=\zeta(\zeta_0, {\bm u}, t)$.

The continuous-time system (\ref{x0}) can be represented as an (infinite) transition system $\mathcal{T}=(X, X_0, \Sigma, \rightarrow, f, O, g)$, where
\begin{itemize}
  \item $X$ is the set of states,
  \item $X_0\subseteq X$ is the set of initial states,
  \item $\Sigma=\mathcal{U}$ is the set of input functions,
  \item $\rightarrow: X\times \Sigma \to  2^X$ is the transition relation,
  \item $O=Y$ is the set of observations, and
  \item $g$ is the observation map.
\end{itemize}
The transition relation $x'\in \rightarrow(x, u)$ if and only if $x'=\xi(x, u, \tau)$, where $\tau>0$ is a given constant. For convenience, $x'\in \rightarrow(x, u)$ will be denoted as $x \xrightarrow[]{u} x'$.
%

\begin{definition}[Controlled transition system]\label{def_cts}
Given a transition system $\mathcal{T}=(X, X_0, \Sigma, \rightarrow, f, O, g)$ and a set of atomic propositions $AP$, we define the controlled transition system (CTS) $\mathcal{T}_c = (X, X_{0}, AP, \rightarrow, L_c)$, where
\begin{itemize}
  \item $L_c: X \to 2^{AP}$ is a labelling function.
\end{itemize}
\end{definition}
The labelling function $L_c(x)$ maps a state $x$ to the finite set of $AP$ which are true at state $x$. Given a state $x\in X$, define
\begin{equation}\label{post}
  Post(x):=\{x'\in X: \exists u\in U, x \xrightarrow[]{u} x'\}.
\end{equation}
An infinite \emph{path} of the CTS $\mathcal{T}_c$ is a sequence of states $\bm{\varrho}=x_0x_1x_2\ldots$ generated by an infinite sequence of inputs ${\bm u}=u_0u_1u_2\ldots$ such that $x_0\in X_0$ and $x_k \xrightarrow[]{u_k} x_{k+1}$ for all $k\ge 0$. Its \emph{trace} is the sequence of atomic propositions that are true in the states along the path, \textit{i.e.,} $\texttt{Trace}(\bm{\varrho})=L_c(x_0)L_c(x_1)L_c(x_2)\ldots$. The satisfaction relation $\bm{\varrho}\models \varphi$ if and only if $\texttt{Trace}(\bm{\varrho})\in \texttt{Words}(\varphi)$.


\subsection{Product automaton and potential functions}

\begin{definition}[Product B\"{u}chi automaton \cite{Baier2008}]\label{productautomaton}
Given a CTS $\mathcal{T}_c = (X, X_{0}, AP, \rightarrow, L_c)$ and a NBA $\mathsf{B}= (S, S_{0}, 2^{AP}, \delta, F)$, the product B\"{u}chi automaton (PBA) is $\mathcal{P}=\mathcal{T}_c \times \mathsf{B}=(S_p, S_{0, p}, 2^{AP}, \delta_p, F_p)$, where $S_p:=X\times S, S_{0, p}:=X_0\times S_{0}, F_p:=(X\times F)\cap S_p$ and
\begin{itemize}
  \item $\delta_p\subseteq S_p \times S_p$, defined by $((x, s), (x', s'))\in \delta_p$ if and only if $x'\in Post(x)$ and $s'\in \texttt{Post}(s)$.
\end{itemize}
\end{definition}

Given a state $p\in S_p$, define the projection operator $\texttt{pj}_{X}(p): S_p \to X$ as a mapping from $p$ to its first component $x\in X$. Given a state $x\in X$, define the function $\beta_{\mathcal{P}}: X\to 2^{S}$, given by
\begin{equation}\label{Buchicorresponds}
  \beta_{\mathcal{P}}(x):=\{s\in S: (x, s)\in S_p\},
\end{equation}
as a mapping from $x$ to the subset of B\"{u}chi states $S$ that correspond to $x$. Denote by $D(p, p')$ the set of all finite runs between state $p\in S_p$ and $p'\in S_p$, \textit{i.e.,}
\begin{eqnarray*}
  &&\hspace{-0.5cm}D(p, p'):=\{p_1p_2\ldots p_n: p_1=p, p'=p_n, \\
 &&\hspace{1.5cm} (p_k, p_{k+1})\in \delta_p, \forall k=1, \cdots, n-1; \forall n\ge 2\}.
\end{eqnarray*}
The state $p'$ is said to be reachable from $p$ if $D(p, p')\neq \emptyset$.  The length of a finite run $\bm{p}=p_1p_2\ldots p_n$ in $\mathcal{P}$, denoted by $Lg(\bm{p})$, is given by
\begin{equation*}
  Lg(\bm{p}):=\sum_{i=1}^{n-1}\|\texttt{pj}_{X}(p_{i+1})-\texttt{pj}_{X}(p_{i})\|.
\end{equation*}
For all $p, p'\in S_p$, the distance between $p$ and $p'$ is defined as follows:
\begin{equation}\label{distancefunction}
  d(p, p')=\begin{cases}
                    \min_{\bm{p}\in D(p, p')} Lg(\bm{p}), & \mbox{if }  D(p, p')\neq \emptyset\\
                    \infty, & \mbox{otherwise}.
                  \end{cases}
\end{equation}

The following definitions of self-reachable set and potential functions are given in \cite{vasile2013sampling}.

\begin{definition}
A set $A\subseteq S_p$ is called \emph{self-reachable} if and only if all states in $A$ can reach a state in $A$, \textit{i.e.,} $\forall p\in A, \exists p'\in A$ such that $D(p, p')\neq \emptyset$.
\end{definition}

\begin{definition}
For a set  $B\subseteq S_p$, a set $C\subseteq B$ is called the \emph{maximal self-reachable set} of $B$ if each self-reachable set $A\subseteq B$ satisfies $A\subseteq C$.
\end{definition}

\begin{definition}[Potential function of states in $\mathcal{P}$]\label{potentialfunctionforproductautomata}
The potential function of a state $p\in S_p$, denoted by $V_{\mathcal{P}}(p)$ is defined as:
\begin{equation*}
  V_{\mathcal{P}}(p)=\begin{cases}
  \min_{p'\in F^*_p}\{d(p, p')\}, & \mbox{if } p\notin  F^*_p\\
  0, & \mbox{otherwise},
\end{cases}
\end{equation*}
where $F^*_p$ is the maximal self-reachable set of the set of accepting states $F_p$ in $\mathcal{P}$ and $d(p, p')$ is defined in (\ref{distancefunction}).
\end{definition}

\begin{definition}[Potential function of states in $\mathcal{T}_c$]\label{potentialfunctionfortransitionsystem}
Let a state $x\in X$ and a set $M_p\subseteq \beta_{\mathcal{P}}(x)$, where $\beta_{\mathcal{P}}(x)$ is defined in (\ref{Buchicorresponds}). The potential function of $x$ with respect to $M_p$, denoted by $V_{\mathcal{T}_c}(x, M_p)$ is defined as
\begin{equation*}
 V_{\mathcal{T}_c}(x, M_p)=\min_{s\in M_p} \{V_{\mathcal{P}}((x, s))\}.
\end{equation*}
\end{definition}

\begin{remark}\label{rem1}
  If $V_{\mathcal{T}_c}(x, M_p)<\infty$, it means that $\exists s\in M_p$ such that starting from $(x, s)$, there exists a run that reaches a self-reachable accepting state of $\mathcal{P}$.
\end{remark}

\section{Problem Formulation}\label{section_pf}

Consider a group of robots moving in a bounded workspace $\mathbb{W} \subset \mathbb{R}^{l}$. The dynamics of robot $i$ is given by
\begin{equation}\label{x}
\dot{x}_i = F_i(x_i, u_i), i\in \mathcal{V},
\end{equation}
where
\begin{equation*}
  x_i:=(p_i, \zeta_i)\in \mathbb{R}^n
\end{equation*}
represents the state of robot $i$, which contains its position state $p_i\in \mathbb{R}^{l}$ and non-position state $\zeta_i\in \mathbb{R}^{n-l}$ (\textit{e.g.,} orientation and/or velocity), and $u_i\in \mathbb{R}^m$ represents the input of robot $i$. The function $F_i: \mathbb{R}^n \times \mathbb{R}^m \to \mathbb{R}^n$ describes the state evolution of robot $i$. The output of robot $i$ is the position state, \textit{i.e.,} $y_i=\texttt{proj}_l(x_i)=p_i, \forall i$.

The state and input of robot $i$ are constrained to the following compact sets
\begin{equation}\label{cons}
  \begin{aligned}
  x_i(t)\in  \mathbb{X}_i, u_i(t)\in  \mathbb{U}_i, \forall t\ge 0.
  \end{aligned}
\end{equation}
It is assumed that the set $\mathbb{U}_i$ contains the origin for all $i$.

Denote by ${\bm \xi}_i: [0, \infty)\to \mathbb{R}^{n}$ the  trajectory of robot $i$ with dynamics given by (\ref{x}). Then, we further define ${\bm p_i}: [0, \infty)\to \mathbb{R}^{l}$ as the position trajectory of robot $i$, where $p_i(t)=\texttt{proj}_{l}({\xi}_i(t)), \forall t\in \texttt{dom}({\bm \xi}_i)$.
Given a time interval $[t_1, t_2], t_1<t_2$, the corresponding trajectory and position trajectory are denoted by ${\bm \xi}_i([t_1, t_2])$ and ${\bm p}_i([t_1, t_2])$, respectively. Denote by ${\bm \xi}_i([t, \infty))$ and ${\bm p}_i([t, \infty))$ the trajectory and the position trajectory of robot $i$ from time $t$ onwards, respectively.

Supposing that the sensing radius of each robot is the same, given by $R>0$, then the communication graph formed by the group of robots is undirected. The neighboring set of robot $i$ at time $t$ is given by $\mathcal{N}_i(t)=\{j\in \mathcal{V}: \|x_i(t)-x_j(t)\|\le R,  j\neq i\}$, so that $j\in \mathcal{N}_i(t)\Leftrightarrow i\in \mathcal{N}_j(t), \forall i\neq j, \forall t$.

The group of robots are working in a common workspace $\mathbb{W}$, which is populated with a set of closed sets $O_i$, corresponding to obstacles. Let $\mathbb{O}={\cup}_{i} O_i$, then the free space $\mathbb{F}$ is denoted by $\mathbb{F}:=\mathbb{W}\setminus \mathbb{O}$.

Each robot $i$ is subject to its own specification $\varphi_i$, which is in the form of a LTL$_{-\mathsf{X}}$ formula that is defined over the workspace $\mathbb{W}$. LTL$_{-\mathsf{X}}$ \cite{emerson1990temporal} is a known fragment of LTL, in which the $\bigcirc$ (``next") operator is not allowed. The choice of LTL$_{-\mathsf{X}}$ over LTL is motivated by the fact that LTL (given in (\ref{LTL})) increases expressivity (over LTL$_{-\mathsf{X}}$) only over words with a finite number of repetitions of a symbol, and a word corresponding to a continuous signal will never have a finite number of successive repetitions of a symbol.

Suppose that a cell decomposition is given over the workspace $\mathbb{W}$. The cell decomposition is a partition of $\mathbb{W}$
into finite disjoint convex regions $\Phi:=\{X_1, \ldots, X_{M}\}$ with $\mathbb{W}=\cup_{l=1}^{M} X_l$. Given a point $p\in \mathbb{W}$, define the map $Q: \mathbb{W}\to \Phi$ as
\begin{equation*}
Q(p):=\{X_l\in\Phi: p\in X_l\},
\end{equation*}
which maps a point $p$ into a cell $X_l\in\Phi$ that contains it. Let $AP_{\varphi_i}$ be the set of atomic propositions specified by $\varphi_i$. Define $AP=\cup_{i\in \mathcal{V}} AP_{\varphi_i}$. Then, we have the following assumption.

\begin{assumption}\label{ass3}
  The cell decomposition over the workspace $\mathbb{W}$ satisfies
\begin{equation*}
L_c(x)=L_c(x'), \forall x, x': Q(x)=Q(x'),
\end{equation*}
where $L_c$ given in Definition \ref{def_cts} is the labelling function.
\end{assumption}

Assumption \ref{ass3} means that for all points that are contained in the same cell, the subset of $AP$ that is true at these points is the same.
We note that the required cell decomposition can be computed exactly or approximately using many existing approaches (Chapters 4-5 \cite{latombe2012robot}).

Given a trajectory ${\bm \xi}_i$, the notation ${\bm \xi}_i\models \varphi_i$ means that the trajectory ${\bm \xi}_i$ satisfies the specification $\varphi_i$. Given the position $p_i$ of robot $i$, we refer to its \emph{footprint} $\phi_i(x_i)$ as the set of points in $\mathbb{W}$ that are occupied by robot $i$ in this position. We note that the footprint of robot $i$ can take into account not only the shape of robot $i$ but also practical issues such as measurement errors.

The objective of this paper is to find, for each robot $i$, a trajectory ${\bm \xi}_i$ such that ${\bm \xi}_i\models \varphi_i$ on the premise that safety (no collisions with static obstacles and no inter-robot collisions) is guaranteed. Let $t=0$ be the task activation time of robot $i, \forall i$. Then, the centralized and offline version of the MRMC problem is formulated below:
\begin{subequations}\label{objec}
\begin{eqnarray}
&&\hspace{-1cm}\text{find} \quad  \{{\bm \xi}_i([0, \infty))\}_{i\in \mathcal{V}}\\
&&\hspace{-2cm}\text{subject to}\nonumber \\
&&\hspace{-1cm} (\ref{x}) \; \text{and} \; (\ref{cons}), \forall i\in \mathcal{V}, \label{c0}\\
&&\hspace{-1cm} {\bm \xi}_i([0, \infty))\models \varphi_i, \forall i\in \mathcal{V}, \label{c0}\\
&&\hspace{-1cm} \phi_i({\bm p}_i([0, \infty))\subset \mathbb{F}, \forall i\in \mathcal{V}, \label{c1}\\
&&\hspace{-1cm} \phi_i({p}_i(t))\cap \phi_j(p_j(t)) = \emptyset, \forall i, j\in \mathcal{V}, i\neq j, \forall t. \label{c2}
\end{eqnarray}
\end{subequations}
Constraint (\ref{c1}) means that the footprint of each robot will not collide with the static obstacles at any time. Constraint (\ref{c2}) means that the footprint of two different robots can not intersect at any time, thus guaranteeing no inter-robot collision occurs. Note that in this paper, each robot has only local view and local information, \textit{i.e.,} each robot considers only robots in its neighborhood $\mathcal{N}_i(t)$ at each time $t$ and can have only local information of its neighbors. Therefore, centralized motion coordination can not be conducted. Under these settings, the MRMC problem (\ref{objec}) is broken into local distributed motion coordination problems and solved online for individual robots. Let ${\bm p}_j([t, t_j^*(t)])$ be the local position trajectory of robot $j$ that is available to robot $i$ at time $t$, where $t_j^*(t):=\min_{t'>t}\{x_j(t')\notin \mathcal{B}(x_i(t), R)\}
$. Then, the (online) motion coordination problem for robot $i$ is formulated as
\begin{subequations}\label{objec2}
\begin{eqnarray}
&&\hspace{-1cm}\text{find} \quad  {\bm \xi}_i([t, \infty))\\
&&\hspace{-1.5cm}\text{subject to}\nonumber \\
&&\hspace{-1.1cm} (\ref{x}) \; \text{and} \; (\ref{cons}),\\
&&\hspace{-1.1cm}  {\bm \xi}_i([0, t] \cup [t, \infty))\models \varphi_i,\\
&&\hspace{-1.1cm} \phi_i({\bm p}_i([t, \infty)))\subset \mathbb{F},  \\
&&\hspace{-1.1cm} \phi_i(p_i(t))\cap \phi_j(p_j(t)) = \emptyset, \forall j\in \mathcal{N}_i(t), \forall t'\in [t, t_j^*(t)],
\end{eqnarray}
\end{subequations}
where ${\bm \xi}_i([0, t])$ is the history trajectory.

\section{Solution}

The proposed solution to the motion coordination problem (\ref{objec2}) consists of three layers: 1) an offline pre-computation layer, 2) an initialization layer, and 3) an online coordination layer.

\subsection{Structure of each robot}

\begin{figure}
  \centering
  \includegraphics[width=.45\textwidth]{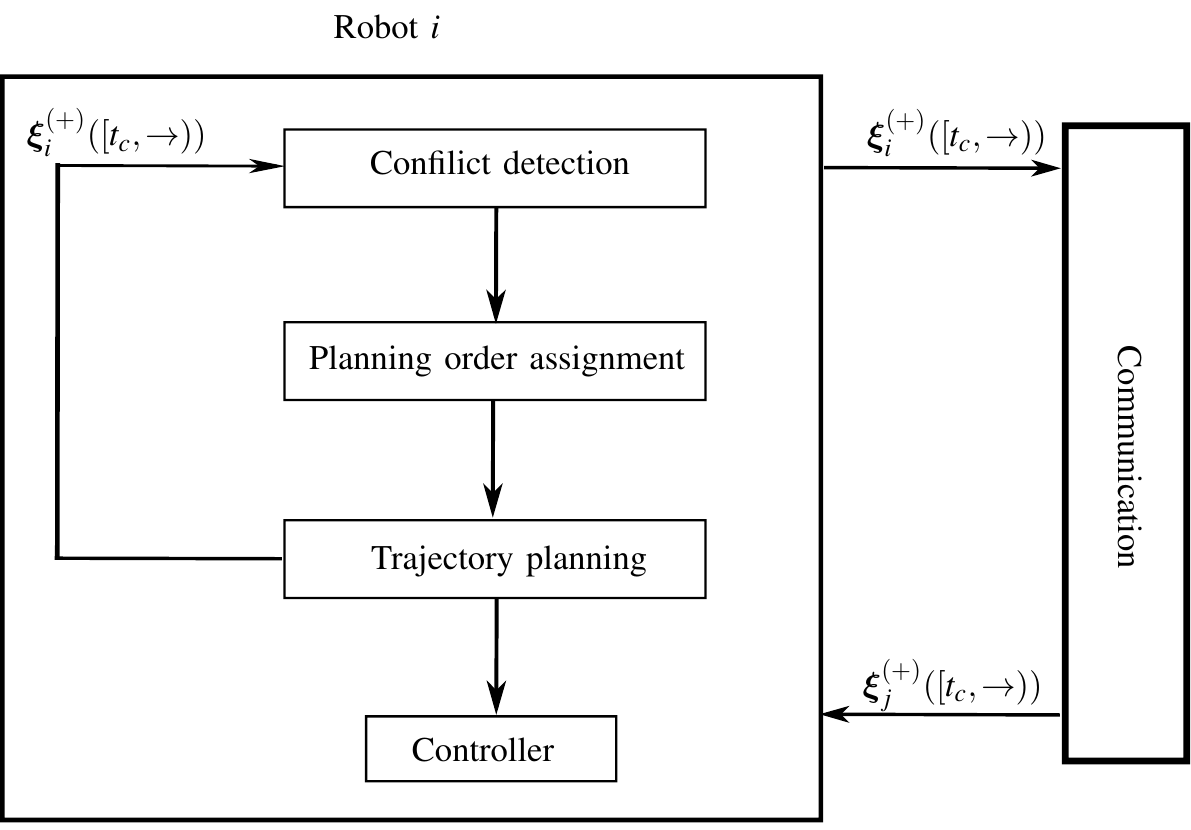}
  \caption{The structure of robot $i$.}\label{fig1}
\end{figure}

Before explaining the solution, the structure of each robot is presented (see Fig. \ref{fig1}). Each robot $i$ is equipped with five modules, the conflict detection, the planning order assignment, the trajectory planning, the control, and the communication module. The first four modules work sequentially while the communication module works in parallel with the first four.

\begin{figure}[H]
  \centering
  \includegraphics[width=.45\textwidth]{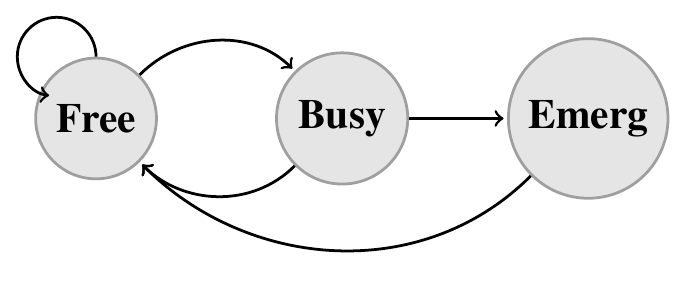}
\caption{The three modes of robot $i$ and the transitions among them.}\label{fig2}
\end{figure}

During online execution, robot $i$ tries to satisfy its specification safely by resolving conflicts with other robots. This is done by following some mode switching rules encoded into a finite state machine (FSM), see Fig. \ref{fig2}. Each FSM has the following three modes:

\begin{itemize}
  \item \textbf{Free}: Robot moves as planned. This is the normal mode.
  \item \textbf{Busy}: Robot enters this mode when conflicts are detected. In this mode, the planning order assignment module and the trajectory planning module are activated.
  \item \textbf{Emerg}: Robot starts an emergency stop process.
\end{itemize}

In Fig. \ref{fig2}, the transitions between different modes of the FSM are depicted. Initially, robot $i$ is in \textbf{Free} mode. The conflict detection module is activated when the online execution starts. Once conflict neighbors (will be defined later) are detected, robot $i$ enters to \textbf{Busy} mode and the planning order assignment and the trajectory planning modules are activated to solve the conflicts, otherwise, robot $i$ stays in \textbf{Free} mode. When robot $i$ is in \textbf{Busy} mode, it switches back to \textbf{Free} mode if the trajectory planning module returns a feasible solution, and the solution will be broadcasted to the robot's neighboring area as well as sent to the controller for execution, otherwise (\textit{e.g.,} no feasible plan is found), robot $i$ switches to \textbf{Emerg} mode and a braking controller (defined later) is applied. Note that when robot $i$ switches to \textbf{Emerg} mode, it will come to a stop but with power-on. This means that robot $i$ will continue monitoring the environment and restart (switches back to \textbf{Free} mode) when it is possible.
%
%
%

\subsection{Offline pre-computation}

\subsubsection{Braking controller}
As stated in the previous subsection, when robot $i$ enters \textbf{Emerg} mode, it starts an emergency stop process. In our previous work \cite{yu2020fully}, safety in \textbf{Emerg} mode is guaranteed by assuming that each robot can take unbounded input when switching to
\textbf{Emerg} mode. In this work, we consider bounded input in all modes for all robots. Due to this, a braking controller needs to be designed for each robot, and the notions of \emph{braking (position) trajectory} and \emph{braking time and distance} are introduced.
%

Given the initial state $x_i\in \mathbb{X}_i$ of robot $i$, define
\begin{subequations}\label{brakingtime}
\begin{eqnarray}
&&\hspace{-1.7cm}t_i^*(x_i):=\min \quad T\\
&&\hspace{-1.5cm}\text{subject to}\nonumber \\
&&\hspace{-1cm} x_i(0)=x_i,\\
&&\hspace{-1cm}  {\dot x}_i(t)=F_i(x_i(t), u_i(t)), \\
&&\hspace{-1cm}  x_i(t)\in \mathbb{X}_i, u_i(t)\in \mathbb{U}_i, t\in [0, T)\\
&&\hspace{-1cm} \dot{p}_i(T)=0_l,
\end{eqnarray}
\end{subequations}
as the minimal time needed to decelerate robot $i$ to zero velocity (\textit{i.e.,} $\dot{p}_i(t)=0_l$). Let ${\bm u}_i^*(x_i)$ be the optimal solution of (\ref{brakingtime}). Then, the \emph{braking controller} ${\bm u}_i^\text{br}(x_i)$ is designed as
\begin{equation}\label{bra_u}
  u_i^\text{br}(x_i)(t)=\begin{cases}
                     {u}_i^*(x_i)(t), & \mbox{if }  \dot{p}_i(t) \neq 0_l,\\
                     0_m, & \mbox{if } \dot{p}_i(t)= 0_l.
                   \end{cases}
\end{equation}
Denote by ${\bm \xi}_i^{\text{br}}(x_i)$ the \emph{braking trajectory} of robot $i$ starting at $x_i$. Then we have
\begin{equation}\label{bra_traj}
\xi_i^{\text{br}}(x_i)(t)=\xi_i(x_i, {\bm u}_i^\text{br}(x_i), t), t\in [0, t_i^*(x_i)].
\end{equation}
The \emph{braking position trajectory} of robot $i$, denoted by ${\bm p}_i^{\text{br}}(x_i)$, is then given by
\begin{equation}\label{bra_postraj}
\begin{aligned}
{\bm p}_i^{\text{br}}(x_i)=\texttt{proj}_{l}({\bm \xi}_i^{\text{br}}(x_i)).
\end{aligned}
\end{equation}
Let $d_i^*(x_i):=\max_{p\in {\bm p}_i^{\text{br}}(x_i)} \|p-\texttt{proj}_l(x_i)\|.$ Then, we have the following assumption.


\begin{assumption}\label{ass2}
Define the \emph{braking time} $T_i^{\text{br}}$ and the \emph{braking distance} $D_i^{\text{br}}$ as
\begin{eqnarray}
  T_i^{\text{br}}&:=&\max_{x_i\in \mathbb{X}_i} \{t_i^*(x_i)\}, \label{safetime}\\
  D_i^{\text{br}}&:=&\max_{x_i\in \mathbb{X}_i}  \{d_i^*(x_i)\}. \label{safedistance}
\end{eqnarray}
Then, we have $0\le T_i^{\text{br}}<\infty$ and $0\le D_i^{\text{br}}<\infty, \forall i$.
\end{assumption}

\begin{remark}
  Assumption \ref{ass2} means that for any given initial state $x_i\in \mathbb{X}_i$, the maximal time and distance needed for braking with the braking controller (\ref{bra_u}) are finite. We note that this assumption is not conservative as it is satisfied by many real-life robots. For robots with first-order dynamics (\textit{e.g.,}  omni-directional robots, differential drive robots), \textit{i.e.,}  the velocity of the robot is controlled, one can derive that $T_i^{\text{br}}=0$ and $D_i^{\text{br}}=0$ with the braking controller $\bm{u}_i^{\text{br}}\equiv 0_m$. For robots with second-order dynamics (\textit{e.g.,}  automated vehicles), \textit{i.e.,} the acceleration of the robot is controlled, the braking time $T_i^{\text{br}}$ and the braking distance $T_i^{\text{br}}$ are usually determined by the maximum velocity and acceleration.
\end{remark}

\begin{example}
Consider a homogeneous MRS, where the dynamics of each robot $i$ is given by
\begin{equation}\label{ex:dynamics}
\begin{aligned}
  \dot{p}_x^i&=v_i\cos\theta_i,\\
  \dot{p}_y^i&=v_i\sin\theta_i,\\
  \dot{\theta}_i&=\omega_i,\\
  \dot{v}_i&=a_i,
  \end{aligned}
\end{equation}
where $p_i:=(p_x^i, p_y^i)\in \mathbb{R}^2$ is the position of robot $i$, $\zeta_i:=(\theta_i, v_i)$ represents the non-position state of robot $i$, which contains its orientation $\theta_i\in \mathbb{R}$ and velocity $v_i\in \mathbb{R}$. The input $u_i$ is given by $u_i:=(\omega_i, a_i)$, where $\omega_i$ is the turning rate and $a_i$ is the acceleration.

The velocity and input of robot $i$ are subject to the hard constraints
\begin{equation}\label{ex:cons}
  \begin{aligned}
  |v_{i}(t)|\le v_{i, \max}, |\omega_{i}(t)|\le \omega_{i, \max}, |a_{i}(t)|\le a_{i, \max},
  \end{aligned}
\end{equation}
where $v_{i, \max}, \omega_{i, \max}, a_{i, \max}>0$. Then, the state and input sets are given by
\begin{eqnarray}
&&\hspace{-1cm} \mathbb{X}_i:=\{(p_i, \theta_i, v_i): p_i\in \mathbb{W}, \theta_i\in \mathbb{R}, |v_{i}|\le v_{i, \max}\}, \label{ex:statecons}\\
&&\hspace{-1cm}  \mathbb{U}_i:=\{(\omega_i, a_i): |\omega_{i}|\le \omega_{i, \max}, |a_{i}|\le a_{i, \max}\}. \label{ex:inputcons}
\end{eqnarray}

In this example, we consider two different braking controllers. First, design the braking controller $\bm{u}_i^{\text{br, 1}}$ as
\begin{equation}\label{brakingcontroller1}
  u_i^\text{br, 1}(t)=\left\{\begin{aligned}
  \left(0, -a_{i, \max}\frac{v_i(t)}{|v_i(t)|}\right), &\quad \text{if}\; |v_i(t)|\neq 0,\\
  0_2 \quad\quad\quad, &\quad \text{if}\; |v_i(t)|= 0.
  \end{aligned}\right.
\end{equation}
Then, one can derive the braking time $T_i^{\text{br, 1}}={|v_{i, \max}|}/{a_{i, \max}}$ and the braking distance $D_i^{\text{br, 1}}=|v_{i, \max}|^2/2{a_{i, \max}}$.

Alternatively, one can design the braking controller $\bm{u}_i^{\text{br, 2}}$ as
\begin{equation}\label{brakingcontroller2}
  u_i^\text{br, 2}(t)=\left\{\begin{aligned}
  \left((-)\omega_{i, \max}, -a_{i, \max}\frac{v_i(t)}{|v_i(t)|}\right), &\quad \text{if}\; |v_i(t)|\neq 0,\\
  0_2 \quad\quad\quad, &\quad \text{if}\; |v_i(t)|= 0.
  \end{aligned}\right.
\end{equation}
Then, one can derive the braking time $T_i^{\text{br, 2}}={|v_{i, \max}|}/{a_{i, \max}}$ and the braking distance \begin{equation*}
D_i^{\text{br, 2}}=\frac{g(v_{i, \max}, \omega_{i, \max}, a_{i, \max})}{\omega_{i, \max}^2},
\end{equation*}
where $g(v_{i, \max}, \omega_{i, \max}, a_{i, \max})=v_{i, \max}^2\omega_{i, \max}^2+2a_{i, \max}^2\Big(1-\cos({v_{i, \max}\omega_{i, \max}}/{a_{i, \max}})\Big)-2v_{i, \max}\omega_{i, \max}a_{i, \max}\sin{({v_{i, \max}\omega_{i, \max}}/{a_{i, \max}})}$.
Note that using the braking controller (\ref{brakingcontroller2}), one can prove that the minimal braking distance is achieved.
\end{example}

\subsubsection{Workspace discretization}

Given the set of $AP$, Assumption \ref{ass3} guarantees that there exists an observation preserving cell decomposition $\Phi=\{X_1, \ldots, X_{M}\}$ over the workspace  $\mathbb{W}$. In general, $X_{l_1}, X_{l_2}, l_1\neq l_2$ are of different shapes and sizes. For the sake of online coordination, we propose to further discretize each cell $X_l\in \Phi$ into smaller regions. Let $\Xi(X_l):=\{\hat X_l^1, \ldots, \hat X_{l}^{M_l}\}$ be a partition of $X_l$ into $M_l$ disjoint convex regions. Then, we define $\Xi:=\cup_{X_l\in \Phi} \Xi(X_l)$. $\Xi$ can be seen as a finer discretization over the workspace  $\mathbb{W}$.

Given a set $S\subseteq \mathbb{W}$, let
\begin{equation}\label{Qp}
\hat Q(S):=\{\hat X\in\Xi: \hat X\cap S \neq \emptyset\},
\end{equation}
which represents the set of regions in $\Xi$ that intersect with $S$.


\subsubsection{Pre-computation of braking area, CTS, NBA, PBA, and potential functions}
Given the position $p_i$ of robot $i$, define
\begin{equation*}
  \mathbb{S}_i(p_i):=\{x_i\in \mathbb{X}_i: \texttt{proj}_l(x_i)=p_i\}
\end{equation*}
as the subset of states in $\mathbb{X}_i$ that correspond to $p_i$. In addition, we define the \emph{braking area} of robot $i$ at position $p_i$ as
\begin{equation}\label{brakingarea}
  \Phi(p_i):=\cup_{x_i\in \mathbb{S}_i(p_i)}\phi_i({\bm p}_i^{\text{br}}(x_i)),
\end{equation}
where $\phi_i({\bm p}_i^{\text{br}}(x_i))$ is the footprint of the braking position trajectory ${\bm p}_i^{\text{br}}(x_i)$. Let
\begin{equation}\label{Re_area}
\psi_i(p_i):=\mathcal{B}(\phi_i(p_i), D_i^{\text{br}}).
\end{equation}
Then, we have the following result.

\begin{proposition}\label{pro1}
The braking area of robot $i$ at position $p_i$ can be over-approximated by the set $\psi_i(p_i)$, that is,
\begin{equation*}
  \Phi(p_i)\subseteq \mathcal{B}(\phi_i(p_i), D_i^{\text{br}}), \forall p_i\in \mathbb{W}.
\end{equation*}
\end{proposition}

In addition, for each region $\hat X_l\in \Xi$, define
\begin{equation}\label{Re_area}
\begin{aligned}
\psi_i(\hat X_l):=&\cup_{p_i\in \hat X_l}\psi_i(p_i).
\end{aligned}
\end{equation}
Each robot $i$ will compute offline a map $M_i: \Xi \to 2^{\Xi}$, where
\begin{equation}\label{map}
\begin{aligned}
M_i(\hat X_l):= \hat Q\left(\psi_i(\hat X_l)\right), \forall \hat X_l\in \Xi.
\end{aligned}
\end{equation}
Intuitively, $M_i$ projects a region $\hat X_l$ into a set of regions that might be traversed by the braking position trajectory of robot $i$ if robot $i$ starts an emergency stop process inside $\hat X_l$.

Due to the continuity of the dynamics, the state and the input spaces, the transition system that represents (\ref{x}) is infinite for each robot $i$. To this end, a probabilistically complete sampling-based algorithm is proposed in \cite{vasile2013sampling} to approximate (\ref{x}) by a finite transition system. Given a sampling interval $\tau_s$, the finite transition system that represents (\ref{x}) is denoted by $\mathcal{T}_i:=(\hat{\mathbb{X}}_i, \hat{\mathbb{X}}_i^0, \mathbb{U}_i, \rightarrow_{c,i}, F_i, \mathbb{W}, h)$, where
\begin{itemize}\label{T}
  \item $\hat{\mathbb{X}}_i$ collects all sampling points in $\mathbb{X}_i$ that are safe (with respect to the static obstacles), \textit{i.e.,} $\psi_i(\texttt{proj}_l(x_i))\subseteq \mathbb{F}, \forall x_i\in \hat{\mathbb{X}}_i$),
  \item $\hat{\mathbb{X}}_i^0\subseteq \hat{\mathbb{X}}_i$,
  \item $\rightarrow_{c,i}\subseteq \hat{\mathbb{X}}_i\times \mathbb{U}_i \times \hat{\mathbb{X}}_i$,
\end{itemize}
$F_i$ is given in (\ref{x}), $\mathbb{W}$ is the workspace as well as the observation space, and $h(\cdot)=\texttt{proj}_l(\cdot)$ is the observation map. Here, $\mathbb{X}_i, \mathbb{U}_i$ are the set of states and inputs, which are defined in (\ref{cons}). The transition relation $(x_i, u_i, x'_i) \in \rightarrow_{c,i}$ if and only if $x'_i=\xi_i(x_i, u_i, \tau_s)$.
Similarly to (\ref{post}), define
 \begin{equation}\label{post2}
   Post(x_i):=\{x'_i\in \hat{\mathbb{X}}_i:  \exists u_i\in \mathbb{U}_i, x_i \xrightarrow[]{u_i} x'_i\}.
 \end{equation}

Once $\mathcal{T}_i$ is obtained, one can further construct the NBA $\mathsf{B}_i:=(S_i, S_i^{0}, 2^{AP_{\varphi_i}}, \delta_i, F_i)$ for the specification $\varphi_i$ (Definition \ref{Buchi}), the CTS $\mathcal{T}_{c, i}:=(\hat{\mathbb{X}}_i, \hat{\mathbb{X}}_i^0, AP_{\varphi_i}, \rightarrow_{c,i}, L_{c,i})$ (Definition \ref{def_cts}), and then form the PBA $\mathcal{P}_i=\mathcal{T}_{c, i} \times \mathsf{B}_i$ (Definition \ref{productautomaton}). After that, the potential function for $\mathcal{P}_i$ can be computed according to Definition \ref{potentialfunctionforproductautomata}.

\begin{remark}
  The cell decomposition $\Phi$ of the workspace satisfies $L_c(x)=L_c(x'), \forall Q(x)=Q(x')$. In addition, $AP_{\varphi_i}\subseteq AP, \forall i\in \mathcal{V}$. Therefore, one has $L_{c, i}(x)=L_{c, i}(x'), \forall Q(x)=Q(x'), \forall i\in \mathcal{V}$ and thus $\beta_{\mathcal{P}_i}(x)=\beta_{\mathcal{P}_i}(x'), \forall Q(x)=Q(x'), \forall i\in \mathcal{V}$. Then, according to Definition \ref{potentialfunctionforproductautomata}, one can get that if $V_{\mathcal{P}_i}((x, \beta_{\mathcal{P}_i}(x)))<\infty$, then $V_{\mathcal{P}_i}((x', \beta_{\mathcal{P}_i}(x')))<\infty, \forall x'\in Q(x)$.
\end{remark}

\subsection{Initialization}

Before proceeding, the following definition is required.
\begin{definition}
  We call a trajectory ${\bm \xi}_i$ of (\ref{x}) \emph{safely satisfy} an LTL formula $\varphi_i$ if i) ${\bm \xi}_i\models \varphi_i$ and ii) $\psi_i(p_i)\subseteq \mathbb{F}, \forall {p}_i\in \texttt{proj}_l({\bm \xi}_i)$.
\end{definition}

At the task activation time $t=0$, robot $i$ first finds a trajectory ${\bm \xi}_i^0$ that safely satisfy $\varphi_i$. The trajectory planning problem for a single robot can be solved by many existing methods, such as search- or sampling-based \cite{bhattacharya2010search,lavalle2001rapidly}, automata-based  \cite{quottrup2004multi}, and optimization-based methods \cite{howard2010receding,schulman2013finding}. We
note that the initial trajectory planning is not
the focus of this paper. For details about this process, we refer to interested readers to corresponding literatures and the references therein.

The following assumption is needed to guarantee the feasibility of each task specification $\varphi_i$.

\begin{assumption}\label{ass1}
  Initially, for each robot $i$, there exists a trajectory ${\bm \xi}_i^0$ that safely satisfy $\varphi_i$.
\end{assumption}

\subsection{Online motion coordination}\label{section_dm}

The initially planned trajectory of each robot does not consider the motion of other robots. Moreover, each robot has only local view and local information (about other robots). Therefore, motion coordination is required during online implementation.
Based on the sensing information (about the workspace) and broadcasted information (from neighboring robots), each robot can detect conflicts within its neighborhood and then conduct motion replanning such that conflicts are avoided. In this work, we consider that the conflict detection is conducted at a sequence of sampling instants $\{t_k\}_{k\in \mathbb{N}}$ for each robot $i$, where $t_{k+1}-t_k\equiv \Delta$.

\subsubsection{Conflict detection}\label{section_cd}

Before proceeding, the following notation is introduced. Given a position trajectory ${\bm p}_i([t_1, t_2])$ and a region $\hat X_l\in \Xi$, the function $\Gamma: \mathcal{F}(\mathbb{R}_{\ge 0}, \mathbb{R}^{p})\times \Xi \to 2^{\mathbb{R}_{\ge 0}}$, defined as
\begin{equation}\label{Ti}
  \Gamma({\bm p}_i([t_1, t_2]), \hat X_l):=\{t\in [t_1, t_2]: p_i(t)\in \hat X_l\},
\end{equation}
gives the time interval that the position trajectory $p_i([t_1, t_2])$ occupies the region $\hat X_l$.


Given the position $p_i(t_k)$, $\mathcal{B}(p_i(t_k), R)$ represents the sensing area of robot $i$ at time $t_k$ and $\hat Q(\mathcal{B}(p_i(t_k), R))$ represents the set of regions in $\Xi$ that intersect with $\mathcal{B}(p_i(t_k), R)$. Let
$$
t_{i}^{fl}(t_k):=\min_{t>t_k}\{p_i(t)\notin \mathcal{B}(p_i(t_k), R)\}
$$
be the first time that robot $i$ leaves its sensing area $\mathcal{B}(p_i(t_k), R)$. Then, define
\begin{equation}\label{Si}
S_i(t_k):= \hat Q\big({\bm p}_i([t_k, t_{i}^{fl}(t_k)])\big)
\end{equation}
as the set of regions traversed by robot $i$ within $\mathcal{B}(p_i(t_k), R)$ until it leaves it at $t_{i}^{fl}(t_k)$. Moreover, for each $\hat X_l\in S_i(t_k)$, the braking area of robot $i$ is contained in $\psi_i(\hat X_l)$. Then, we define
\begin{equation}\label{resi}
Res_i(\hat X_l):=M_i(\hat X_l)
\end{equation}
as the set of \emph{reserved regions} by robot $i$ in order to safely brake when robot $i$ is within $\hat X_l$. According to (\ref{Ti}), the time interval that robot $i$ occupies the region $\hat X_l\in S_i(t_k)$ is given by $\Gamma({\bm p}_i([t_k, t_{i}^{fl}(t_k)]), {\hat X}_l)$. In addition, by the continuity of ${\bm p}_i([t_k, \infty))$, one can conclude that $\Gamma({\bm p}_i([t_k, t_{i}^{fl}(t_k)]), {\hat X}_l)$ is given by one or several disjoint time interval(s) of the form $[a, b), a<b$. Supposing that \begin{equation}\label{traj_time}
  \Gamma({\bm p}_i([t_k, t_{i}^{fl}(t_k)]), {\hat X}_l)=\cup_{l=1}^m [a_l, b_l),
\end{equation}
where $m$ is the number of disjoint intervals in $\Gamma({\bm p}_i([t_k, t_{i}^{fl}(t_k)]), {\hat X}_l)$. Denote by $\mathcal{T}_i(\hat X_l)$ the time interval that robot $i$ reserves the area $Res_i(\hat X_l)$. Then, it can be over-approximated by
\begin{equation}\label{res_time}
\mathcal{T}_i(\hat X_l):=\cup_{l=1}^{m}[a_l, b_l+T_i^{\text{br}}),
\end{equation}
where $T_i^{\text{br}}$ is defined in (\ref{safetime}). Then, we have the following definitions.

\begin{definition}\label{def2}
We say that there is a \emph{spatial-temporal conflict} between robot $i$ and $j$ at time $t_k$ if $\exists \hat X_{l}\in S_i(t_k), \hat X_{l'}\in S_j(t_k)$ such that $Res_i(\hat X_l)\cap  Res_j(\hat X_{l'})\neq \emptyset$ and $\mathcal{T}_i(\hat X_l)\cap  \mathcal{T}_j(\hat X_{l'})\neq \emptyset$.
\end{definition}

Based on Definition \ref{def2}, we define the set of conflict neighbors of robot $i$ at time $t_k$, denoted by $\tilde{\mathcal{N}}_i(t_k)$, as
\begin{equation}\label{conflictneighbor}
\begin{aligned}
\tilde{\mathcal{N}}_i&(t_k):=\{j\in \mathcal{N}_i(t_k):\exists \hat X_{l}\in S_i(t_k), \hat X_{l'}\in S_j(t_k) \;\text{s.t.}\;\\
& Res_i(\hat X_l)\cap Res_j(\hat X_{l'})\neq \emptyset \wedge \mathcal{T}_i(\hat X_l)\cap \mathcal{T}_j(\hat X_{l'})\neq \emptyset\}.
\end{aligned}
\end{equation}

Then, we have the following Proposition.

\begin{proposition}\label{pro2}
For robot $i$, if $\tilde{\mathcal{N}}_i(t_k)=\emptyset$, then one has
\begin{itemize}
  \item [i)] $\psi_i(p_i(t))\cap \psi_j(p_j(t)) = \emptyset, \forall j\in \mathcal{N}_i(t_k), \forall t\in [t_k, t_{i}^{fl}(t_k)]$;
  \item [ii)] $\phi_i(p_i(t))\cap \phi_j(p_j(t)) = \emptyset, \forall j\in \mathcal{N}_i(t_k), \forall t\in [t_k, t_{i}^{fl}(t_k)]$.
\end{itemize}
\end{proposition}

\textbf{\emph{Proof:}} Since $T_i^{\text{br}}$ is the maximum time required to decelerate robot $i$ to zero velocity under the braking controller (\ref{bra_u}), one has that $\Gamma({\bm p}_i^{\text{br}}(p_i(t)), {\hat X}_l)\subseteq \mathcal{T}_i(\hat X_l), \forall t\in [t_k, t_{i}^{fl}(t_k)]$. In addition, according to (\ref{Re_area}), (\ref{map}), and (\ref{resi}), one has that $\psi_i(p_i(t))\subseteq Res_i(\hat X_l), \forall p_i(t)\in \hat X_l$. That is to say, $\tilde{\mathcal{N}}_i(t_k)=\emptyset$ implies i).

For each region $\hat X_l\in \Xi$, define
\begin{equation*}
\phi_i(\hat X_l):=\cup_{p_i\in \hat X_l}\phi_i(p_i).
\end{equation*}
Then, one has $\phi_i(\hat X_l)\subseteq \psi_i(\hat X_l)$. The time interval that the footprint of robot $i$ occupies $\hat X_l$ is given by $\Gamma({\bm p}_i([t, t_i^{fl}(t_k)]), {\hat X}_l)$ and $\Gamma({\bm p}_i([t, t_i^{fl}(t_k)]), {\hat X}_l)\subseteq \mathcal{T}_i(\hat X_l)$. Thus, one can further get that i) implies ii).
$\square$


Robot $i$ switches to \textbf{Busy} mode if and only if the set of conflict neighbors is non-empty (\textit{i.e.,} $\tilde{\mathcal{N}}_i(t_k)\neq \emptyset$). The conflict detection process is outlined in Algorithm 1.

\begin{algorithm}\label{algorithm1}
\caption{\textit{conflictDetection}}
\begin{algorithmic}[1]
\Require $S_j(t_k), \Gamma({\bm p}_j([t_k, t_j^{fl}(t_k)]), {\hat X}_l), \forall \hat X_l\in S_j(t_k)$ for each $j\in \mathcal{N}_i^+(t_k)$.
\Ensure Set of conflict neighbors $\tilde{\mathcal{N}}_i(t_k)$.
\State Initialize $\tilde{\mathcal{N}}_i(t_k)=\emptyset$.
\State Compute $Res_j(\hat X_l), \mathcal{T}_j(\hat X_l)$ for each $j\in \mathcal{N}_i^+(t_k), \hat X_l\in S_j(t_k)$,
\For {$j\in \mathcal{N}_i(t_k)$}
\If {$\exists \hat X_{l}\in S_i(t_k), \hat X_{l'}\in S_j(t_k)$ s.t. $
 Res_i(\hat X_l)\cap Res_j(\hat X_{l'})\neq \emptyset \wedge \mathcal{T}_i(\hat X_l)\cap \mathcal{T}_i(\hat X_{l'})\neq \emptyset$}
\State $\tilde{\mathcal{N}}_i(t_k)=\tilde{\mathcal{N}}_i(t_k)\cup \{j\}$,
\EndIf
\EndFor
\end{algorithmic}
\end{algorithm}

\begin{remark}
To implement Algorithm 1, each robot $i$ needs
only to broadcast to its neighboring area local information
about its plan. To be more specific, which region (\textit{e.g.,} $\hat X_l$) within the sensing area $\mathcal{B}(p_i(t_k), R)$ is occupied by
robot $i$ and when that happens (\textit{i.e.,} $\Gamma({\bm p}_i([t, t_i^{fl}(t_k)]), {\hat X}_l)$). Note that the non-position information (\textit{i.e.,} $\zeta_i(t)$) of each robot is not required to be broadcasted to the neighbors.
\end{remark}

\subsubsection{Determine planning order}\label{section_pa}

Based on the neighboring relation and conflict relation, the graph $\mathcal{G}(t_k)=\{\mathcal {V}, \mathcal {E}(t_k)\}$ formed by the group of robots is naturally divided into one or multiple connected subgraphs, and the motion planning is conducted in parallel within each subgraph in a sequential manner. In order to do that, a planning order needs to be decided for each connected subgraph. In this work, we propose a simple rule to assign priorities between each pair of neighbors.


The number of neighbors and conflict neighbors of robot $i$ at time $t_k$ are given by $|\mathcal{N}_i(t_k)|$ and $|\tilde{\mathcal{N}}_i(t_k)|$, respectively. Then, we have the following definition.

\begin{definition}\label{def3}
We say that robot $i$ has \emph{advantage} over robot $j$ at time $t_k$ if $\tilde{\mathcal{N}}_j(t_k)\neq \emptyset$ and

1) $|\mathcal{N}_i(t_k)|>|\mathcal{N}_j(t_k)|$; OR

2) $|\mathcal{N}_i(t_k)|= |\mathcal{N}_j(t_k)|$ and $|\tilde{\mathcal{N}}_i(t_k)|>|\tilde{\mathcal{N}}_j(t_k)|$.
\end{definition}

Let $\mathcal{Y}_i(t_k)$ be the set of neighbors that have higher priority than robot $i$ at time $t_k$. The planning order assignment process is outlined in Algorithm 2.

For each neighbor $j\in \mathcal{N}_i(t_k)$, if $j$ is in \textbf{Emerg} mode (and thus will be viewed as a static obstacle) or $\tilde{\mathcal{N}}_j(t_k)=\emptyset$, then robot $j$ has higher priority (lines 3-5). Otherwise, robot $j$ has higher priority in motion planning if robot $j$ has advantage over robot $i$ (lines 6-8). However, for the special case, \textit{i.e.,} $|\mathcal{N}_i(t_k)|= |\mathcal{N}_j(t_k)|$ and $|\tilde{\mathcal{N}}_i(t_k)|=|\tilde{\mathcal{N}}_j(t_k)|$, neither robot $i$ nor $j$ has advantage over the other. In this case, the priority is determined by the initially uniquely assigned priority score for each robot $i$ (\textit{i.e.,} $P_i^0\neq P_j^0, \forall i, j$). Denote by $P_i^0$ the priority score of robot $i$. We say that robot $i$ has \emph{priority} over $j$ if $P_i^0>P_j^0$ (lines 9-11).

\begin{algorithm}\label{algorithm2}
\caption{\textit{planningOrderAssignment}}
\begin{algorithmic}[1]
\Require $\mathcal{N}_j(t_k), \tilde{\mathcal{N}}_j(t_k), P_j^0, j\in \mathcal{N}_i(t_k)\cup\{i\}$.
\Ensure Set of higher priority neighbors $\mathcal{Y}_i(t_k)$.
\State Initialize $\mathcal{Y}_i(t_k)=\emptyset$.
\For {$j \in \mathcal{N}_i(t_k)$},
\If {$j$ is in \textbf{Emerg} mode or $\tilde{\mathcal{N}}_j(t_k)=\emptyset$},
\State $\mathcal{Y}_i(t_k)=\mathcal{Y}_i(t_k)\cup j$,
\Else
\If {$j$ has advantage over $i$},
\State $\mathcal{Y}_i(t_k)=\mathcal{Y}_i(t_k)\cup j$,
\Else
\If {neither robot $i$ nor $j$ has advantage over the other and $P_j^0>P_i^0$},
\State $\mathcal{Y}_i(t_k)=\mathcal{Y}_i(t_k)\cup j$,
\EndIf
\EndIf
\EndIf
\EndFor
\end{algorithmic}
\end{algorithm}

\begin{proposition}[Deadlock-free in planning order assignment]
The planning order assignment rule given in Algorithm 2 will result in no cycles, \textit{i.e.,} $\nexists \{q_m\}_1^{\hat K}, {\hat K}\ge 2$ such that $q_{\hat K}\in \mathcal{Y}_{q_1}(t_k)$ and $q_{m-1}\in \mathcal{Y}_{q_m}(t_k), \forall m=2, \ldots, {\hat K}$.
\end{proposition}

\begin{remark}
The rationale behind the rule can be explained as follows. Since the motion planning is conducted sequentially within each subgraph based on the priority order obtained in Algorithm 2, then the total time required to complete the motion planning is given by $K\mathcal{O}(dt)$, where $\mathcal{O}(dt)$ represents the time complexity of one round of motion planning and $K$ represents the number of rounds (if multiple robots conduct motion planning in parallel, it is counted as one round), which is determined by the priority assignment rule being used (\textit{e.g.,} if fixed priority is used, the number of rounds is $K=N$). In our rule, we assign the robot with more neighbors or more conflict neighbors the higher priority, and in this way, we try to minimize the number of rounds needed.
\end{remark}

\begin{figure}
  \centering
  \includegraphics[width=.2\textwidth]{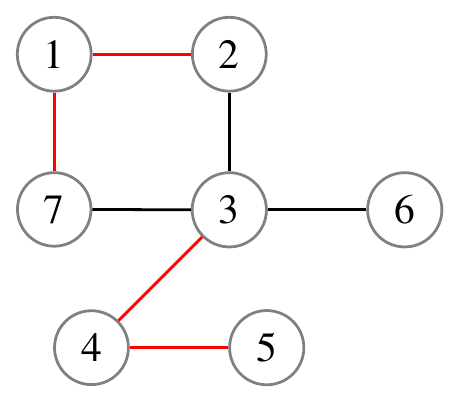}
  \caption{Communication and conflict graph, where both the black and red lines represent communication relation and the red lines represent conflict relation.}\label{fig3}
\end{figure}

\begin{example}
  Consider a group of 7 robots, whose communication relation and conflict relation (at time $t_k$) are depicted in Fig. \ref{fig3}. According to the proposed planning order assignment rule, the motion planning can be completed in 3 rounds, where robots 1 and 3 are in the first round, robots 2, 4 and 7 are in the second round, and robot 5 is in the third round. Note that no motion planning is required for robot 6 since robot 6 has no conflict neighbor.
\end{example}

\subsection{Motion planning}

Before starting to plan, robot $i$ needs to wait for the updated plan from the set of neighbors that have higher priority than robot $i$ (\textit{i.e.,} $j\in \mathcal{Y}_i(t_k)$) and consider them as moving obstacles. For those neighbors $j\in \mathcal{Y}_i(t_k)$, denote by ${\bm \xi}_j^+([t_k, \infty))$ (correspondingly ${\bm \xi}_j^+([t_k, \infty))$) the updated trajectory (position trajectory) of robot $j$ at time $t_k$ and let $t_{j}^{fl+}(t_k)$ be the first time that robot $j$ leaves its sensing area $\mathcal{B}(p_j(t_k), R)$ according to the updated position trajectory ${\bm p}_j^+([t_k, \infty))$. Then, similar to (\ref{Si}), one can define
\begin{equation*}
\begin{aligned}
  S_j^+(t_k):=\hat Q({\bm p}_j^+([t_k, t_{j}^{fl+}]))
  \end{aligned}
\end{equation*}
as the updated set of regions traversed by robot $j$ within $\mathcal{B}(p_j(t_k), R)$. For each $\hat X_l\in S_j^+(t_k)$, the time interval that robot $j$ occupies the region $\hat X_l$ is given by $\Gamma({\bm p}_j^+([t_k, t_{j}^{fl+}]), {\Hat X}_l)$. Supposing that $\Gamma({\bm p}_j^+([t_k, t_{j}^{fl+}]), {\Hat X}_l)=\cup_{l=1}^{\hat m} [\hat a_l, \hat b_l)$, where $\hat m$ is the number of disjoint intervals in $\Gamma({\bm p}_j^+([t_k, t_{j}^{fl+}]), {\Hat X}_l)$. Denote by $\mathcal{T}_j^+(\hat X_l)$ the time interval that robot $j$ reserves the area $Res_j(\hat X_l)$ according to the updated position trajectory ${\bm p}_j^+([t_k, \infty))$. Similarly to (\ref{res_time}), it can be over-approximated by $\mathcal{T}_j^+(\hat X_l):=\cup_{l=1}^{\hat m}[\hat a_l, \hat b_l+T_i^{\text{br}}].$

Then, the trajectory planning problem (TPP) can be formulated as follows:
\begin{subequations}\label{optim}
\begin{eqnarray}
&&\hspace{-0cm}\text{find} \quad {{\bm \xi}_i}([t_k, \infty)),\\
&&\hspace{-1.2cm}\text{subject to}\nonumber \\
&&\hspace{-1cm} (\ref{x})\; \text{and} \; (\ref{cons}),\\
&&\hspace{-1cm} {\bm \xi}_i([0, t_k] \cup [t_k, \infty))\models \varphi_i, \label{optim-c}\\
&&\hspace{-1cm} \psi_i(\texttt{proj}_l({\bm \xi}_i([t_k, \infty))))\subset \mathbb{F}, \label{optim-a}\\
&&\hspace{-1cm} \begin{aligned} \psi_i(\texttt{proj}_l(\xi_i(t)))&\cap Res_j(\hat X_l)=\emptyset,\;t\in \mathcal{T}_{j}^+(\hat X_l), \\
& \forall j\in \mathcal{Y}_i(t_k), \forall \hat X_l\in S_j^+(t_k). \end{aligned} \label{optim-b}
\end{eqnarray}
\end{subequations}
Constraints (\ref{optim-a}) and (\ref{optim-b}) guarantee respectively that there will be no robot-obstacle collisions and inter-robot collisions for robot $i$.

When relatively simple specifications, \textit{e.g.,} reach-avoid type of tasks, are considered, various existing optimization toolboxes, \textit{e.g.,} IPOPT \cite{biegler2009large}, ICLOCS2 \cite{nie2018iclocs2}, and algorithms, \textit{e.g.,} the configuration space-time search \cite{parsons1990motion}, the Hamilton-Jacobian reachability-based trajectory planning \cite{chen2018robust}, RRT$^X$ \cite{otte2016rrtx}, and the fast robot motion planner \cite{lin2018fast} can be utilized to solve (\ref{optim}). However, if the specifications are complex LTL formulas, the constraint (\ref{optim-c}) is not easy to be verified online using the methods mentioned above (the PBA $\mathcal{P}_i$ can be used to verify (\ref{optim-c}), however, it can not deal with the spatial-temporal collision avoidance constraint (\ref{optim-b}) at the same time). Recently, an online RRT-based algorithm is proposed in \cite{vasile2014reactive} to generate local paths that guarantee the satisfaction of the global specification. Motivated by this work, an online motion replanning structure is proposed in this paper, which contains a local and a global trajectory generation algorithms.

Before proceeding, the following notations are required. Denote by
\begin{eqnarray}
 \nonumber 
  && NI_i(t_k):=\{Res_j(\hat X_l), \mathcal{T}_j^+(\hat X_l), \\
  && \hspace{2.6cm} \forall \hat X_l\in S_j^+(t_k), \forall j\in \mathcal{Y}_i(t_k)\}
\end{eqnarray}
the set with respect to robot $i$ which contains all local trajectory information of higher priority neighbors at time $t_k$. Given a state $s_i\in \hat{\mathbb{X}}_i$, define the function $B_i: \hat{\mathbb{X}}_i \to 2^{S_i}$ as a map from a state $s_i\in \hat{\mathbb{X}}_i$ to a subset of valid B\"{u}chi states which correspond to $s_i$ (\textit{i.e.,} $B_i(s_i))\subseteq \beta_{\mathcal{P}_i}(s_i)$, where $\beta_{\mathcal{P}_i}(s_i)$ is defined in (\ref{Buchicorresponds})). Function $B_i$ is used to capture the fact that given a partial trajectory, not all B\"{u}chi states in $\beta_{\mathcal{P}_i}$ are valid. At the task activation time $0$, one has $B_i(\xi_i(0))=\beta_{\mathcal{P}_i}(\xi_i(0))$. During the online implementation, $B_i(\xi_i(t_k)), t_k>0$ is updated by
\begin{equation}\label{Bi}
\begin{aligned}
  B_i(\xi_i(t_k))= \beta_{\mathcal{P}_i}&(\xi_i(t_k)) \\
  &\cap \{B_i(\xi_i(t_{k-1}))\cup \verb"Post"(B_i(\xi_i(t_{k-1})))\}
  \end{aligned}
\end{equation}
where
\begin{equation}\label{PostBi}
  \texttt{Post}(B_i(\xi_i(t_{k-1})))=\cup_{s_i\in B_i(\xi_i(t_{k-1}))} \texttt{Post}(s_i).
\end{equation}

Firstly, the local trajectory generation process is outlined in Algorithm 3. At each time instant $t_k$, Algorithm 3 takes the state $\xi_i(t_k)$ of robot $i$, $B_i(\xi_i(t_k)), \texttt{Post}(B_i(\xi_i(t_k)))$, the offline computed PBA $\mathcal{P}_i$, potential function $V_{\mathcal{P}_{i}}$, the set of static obstacles $\mathbb{O}$, and the local trajectory information of higher priority neighbors, \textit{i.e.,} $NI_i(t_k)$ as input.
The output is a local CTS $\mathcal{T}_{c,i}^L:=(S_i^{L}, S_{i,0}^{L}, AP_{\varphi_i}, \rightarrow_{c,i}^L, L_{c, i})$ that is constructed incrementally and a leaf node $\xi_i^f$.
\begin{algorithm}\label{algorithm3}
\caption{\textit{localTrajectoryGeneration}}
\begin{algorithmic}[1]
\Require $\xi_i(t_k), B_i(\xi_i(t_k)), \texttt{Post}(B_i(\xi_i(t_k)))$, $\mathcal{P}_i, V_{\mathcal{P}_{i}}$, $\mathbb{O}$, and
$NI_i(t_k)$.
\Ensure A local transition system $\mathcal{T}_{c, i}^{L}$ and a leaf node $\xi_i^f$.
\State Initialize $\mathcal{T}_{c, i}^{L}=(S_i^{L}, S_{i,0}^{L}, AP_{\varphi_i}, \rightarrow_{c,i}^L, L_{c, i})$ and $\xi_i^f=\emptyset$, where $S_i^{L} =S_{i,0}^{L}= \xi_i(t_k)$ and $\rightarrow_{c,i}^L =\emptyset$, $\verb"St"(\xi_i(t_k))=0$.
\For {$k=1,\ldots, N_i^{\max}$},
\State $\xi_s\leftarrow \textit{generateSample}(SA_i(t_k))$,
\State $\xi_n\leftarrow \textit{nearest}(S_i^{L}, \xi_s)$,
\State Solve the optimization program $\mathcal{P}(\xi_n, \xi_s, \tau_s)$, which returns $(\xi_r, u_i^*)$,
\State $B_i(\xi_r)\leftarrow \beta_{\mathcal{P}_i}(\xi_r) \cap \{B_i(\xi_n)\cup \verb"Post"(B_i(\xi_n))\}$,
\If {$B_i(\xi_r)\neq \emptyset \wedge V_{\mathcal{T}_{c,i}}(\xi_r, B_i(\xi_r))<\infty$},
\State $\mathbb{O}_i(t_k)\leftarrow\textit{updateObstacle}(\mathbb{O}, NI_i(t_k), [\texttt{St}(\xi_n)\tau_s,$ $(\texttt{St}(\xi_n)+1)\tau_s])$,
\If {$\texttt{dist}(\texttt{proj}_l([\xi_n, \xi_r]), \mathbb{O}_i(t_k))\ge D_i^{\text{br}}$},
\State $S_i^{L}\leftarrow S_i^{L}\cup\{\xi_r\}; \rightarrow_{c,i}^{L}= \rightarrow_{c,i}^{L}\cup \{\xi_n \xrightarrow[]{u_i^*} \xi_r\}$,
\State $\verb"St"(\xi_s)\leftarrow \verb"St"(\xi_n)+1$,
\EndIf
\EndIf
\If {$\texttt{proj}_l(\xi_r)\notin \mathcal{B}(\texttt{proj}_l(\xi_i(t_k)), R)$,}
\State $k=N_i^{\max}+1$,
\State $\xi_i^f\leftarrow \xi_r$,
\EndIf
\EndFor
\end{algorithmic}
\end{algorithm}

In line 5, the optimization program $\mathcal{P}(\xi_n, \xi_s, \tau_s)$ is given by
\begin{subequations}\label{ustar}
\begin{eqnarray}
&&\hspace{-0cm}\min_{u_i\in \mathbb{U}_i} \quad \|\xi_r-\xi_s\|,\\
&&\hspace{-1.2cm}\text{subject to}\nonumber \\
&&\hspace{-1cm} \xi_i(0)=\xi_n,\\
&&\hspace{-1cm} \xi_n+\int_{0}^{t} F_i(\xi_i(s), u_i)ds\in \mathbb{X}_i, \forall t\in [0, \tau_s], \\
&&\hspace{-1cm} \xi_r=\xi_n+\int_{0}^{\tau_s} F_i(\xi_i(s), u_i)ds,
\end{eqnarray}
\end{subequations}
and the optimal solution is $u_i^*$.

The root state of $\mathcal{T}_{c,i}^L$ is robot $i$'s state $\xi_i(t_k)$. The function $\texttt{st}: S_i^L \to \mathbb{N}$ maps a state $x\in S_i^L$ to the number of time steps needed to reach the root state $\xi_i(t_k)$. Initially, $\mathcal{T}_{c,i}^L$ contains one state $\xi_i(t_k)$ and 0 transitions, \textit{i.e.,} $\verb"st"(\xi_i(t_k))=0$, and the leaf node $\xi_i^f=\emptyset$ (line 1). In each iteration (lines 2-18), a new state $\xi_s$ is generated randomly from the set $SA_i(t_k)$ using the $\textit{generateSample}$ procedure (line 3), where $SA_i(t_k)$ is the sampling area around robot $i$, given by
\begin{eqnarray}\nonumber
  &&\hspace{-1cm}SA_i(t_k):=\{(p, \zeta)\in \mathbb{X}_i: p\in \mathcal{B}(\texttt{proj}_l(\xi_i(t_k)), R+\eta)\},
\end{eqnarray}
where $\eta>0$ is an offline chosen constant, which guarantees that there exists $s\in SA_i(t_k)$ such that $\texttt{proj}_l(s)\notin \mathcal{B}(\texttt{proj}_l(\xi_i(t_k)), R)$. This condition is essential for checking the terminal condition (line 15). The $\textit{nearest}$ function (line 4) is a standard RRT primitive \cite{lavalle2001rapidly} which returns the nearest state in $S_i^L$ to the new sample $\xi_s$. Then, one further finds, within the set of states that are reachable from $\xi_n$ at time $\tau_s$, the closest one to the new sample $\xi_s$, \textit{i.e.,} $\xi_r$, and the corresponding input $u_i^*$ (line 5). Here, $\tau_s$ is the sampling interval (the same one used for constructing the transition system $\mathcal{T}_{i}$ in Section \ref{T}). Once $\xi_r$ is obtained, we further compute the subset of valid B\"{u}chi states which correspond to $\xi_r$, \textit{i.e.,} $B_i(\xi_r)$, according to (\ref{Bi}) (line 6). After that, if both conditions $B_i(\xi_r)\neq \emptyset$ and $V_{\mathcal{T}_{c,i}}(\xi_r, B_i(\xi_r))<\infty$ are satisfied (which guarantees that there exists a path, starting from $\xi_r$, that reaches a self-reachable accepting state of $\mathcal{P}_i$, recall Remark \ref{rem1}), obstacles that appear during the time interval $[\texttt{st}(\xi_n)\tau_s, (\texttt{st}(\xi_n)+1)\tau_s]$ are added into the workspace using the function $\textit{updateObstacles}$ (Algorithm 4). Finally, the state $\xi_r$ is added into $S_i^L$ and the transition relation $\xi_n \xrightarrow[]{u_i^*} \xi_r$ is added into $\rightarrow_{c,i}^L$ if the distance between the line segment $\texttt{proj}_l([\xi_n, \xi_r])$ and the obstacles is no less than the braking distance of robot $i$  (lines 9-10), and then the time step needed for $\xi_r$ to reach the root state $\xi_i(t_k)$ is recorded (line 11). The algorithm is terminated when the local sampling tree reaches the outside of the sensing area of robot $i$, and the leaf node $\xi_i^f$ is then given by the corresponding state $\xi_r$ (line 14-17).

\begin{algorithm}\label{algorithm4}
\caption{\textit{updateObstacle}}
\begin{algorithmic}[1]
\Require $\mathbb{O}$, $NI_i(t_k)$ and a time interval $[t_1, t_2]$.
\Ensure $\mathbb{O}_i(t_k)$.
\State $\mathbb{O}_i(t_k)\leftarrow \mathbb{O}$,
\For {$j\in \mathcal{Y}_i(t_k)$,}
\For {$\hat X_l\in S_j^+(t_k)$,}
\If {$\mathcal{T}_j^+(\hat X_l)\cap [t_k+t_1, t_k+t_2]\neq \emptyset$,}
\State $\mathbb{O}_i(t_k)\leftarrow \mathbb{O}_i(t_k)\cup (Res_j(\hat X_l)\cap SA_i(t_k))$,
\EndIf
\EndFor
\EndFor
\end{algorithmic}
\end{algorithm}

\begin{remark}
The complexity of one iteration of Algorithm 3 is the same as for the standard RRT. The functions \textit{generateSample} and \textit{nearest} are standard RRT primitives (one can refer to \cite{lavalle2001rapidly} for more details). The complexity of the \textit{updateObstacle} process (Algorithm 4) at time $t_k$ is $\mathcal{O}(1)$ since $|\mathcal{Y}_i(t_k)|\le N-1$ and $|S_j^+(t_k)|\le |\hat Q(\mathcal{B}(p_j(t_k), R))|, \forall j$, where $|\hat Q(\mathcal{B}(p_j(t_k), R))|$ represents the number of regions contained in the sensing area $\mathcal{B}(p_j(t_k), R)$. The computation of $\texttt{dist}(\texttt{proj}_l([\xi_n, \xi_r]), \mathbb{O}_i(t_k))$ can be formulated as a convex optimization problem and solved in $\mathcal{O}(1)$ since there is a limited number of obstacles in $SA_i(t_k)$ and each obstacle is of the form of a convex region. Moreover, the calculations of $B_i(\xi_r)$ and $V_{\mathcal{T}_{c,i}}(\xi_r, B_i(\xi_r))$ are of the complexity of $\mathcal{O}(1)$ since $\mathcal{P}_{i}, V_{\mathcal{P}_{i}}$ are computed offline.
\end{remark}

After the local CTS $\mathcal{T}_{c,i}^L$ is obtained, we further need to find a path, starting from the leaf node $\xi_i^f$, that reaches one of the maximal self-reachable accepting states $F_{p, i}^*$ of $\mathcal{P}_i$. Define
\begin{equation*}
  P_i(\xi_i^f):=\cup_{s_i\in B_i(\mathcal{T}_{c,i}^L)}(\xi_i^f, s_i)
\end{equation*}
as the set of states in the PBA $\mathcal{P}_i$ that correspond to $\xi_i^f$. Then, the global trajectory generation process is outlined in Algorithm 5. Algorithm 5 takes the set $P_i(\xi_i^f)$ and the the PBA $\mathcal{P}_i$ as input. It first finds the state $p_i^*$ in $P_i(\xi_i^f)$ that has the minimum potential (line 1). Then, if $p_i^*\in F_{p, i}^*$, the function \textit{DijksCycle}($\mathcal{P}_i$, \emph{source}) (defined in \cite{guo2015multi}) is used to compute a shortest cycle from the \emph{source} state back to itself (line 3); otherwise, the function \textit{DijksTargets}($\mathcal{P}_i$, \emph{source}, \emph{targets}) (defined in \cite{guo2015multi}) is used to compute a shortest path in $\mathcal{P}_i$ from ``\emph{source}" state to one of the state belonging to the set ``\emph{targets}" (line 5).
The required path is then the projection of $\bm{p}_i$ on the state space of $\mathcal{T}_{c,i}$ (line 7).

\begin{algorithm}\label{algorithm5}
\caption{\textit{globalTrajectoryGenration}}
\begin{algorithmic}[1]
\Require $P_i(\xi_i^f)$ and $\mathcal{P}_i$.
\Ensure a path $\bm{\rho}_i$.
\State $p_i^*=\min_{p_i\in P_i(\xi_i^f)}\{V_{\mathcal{P}_i}(p_i)\},$
\If {$p_i^*\in F_{p, i}^*$}
\State $\bm{p}_i\leftarrow \textit{DijksCycle}(\mathcal{P}_i, p_i^*)$,
\Else
\State $\bm{p}_i\leftarrow \textit{DijksTargets}(\mathcal{P}_i, p_i^*,F_{p, i}^*)$,
\EndIf
\State $\bm{\rho}_i=\texttt{pj}_{\hat{\mathbb{X}}_i}(\bm{p}_i)$,
\end{algorithmic}
\end{algorithm}

It is possible that after the maximum number of iterations (\textit{i.e.,} $N_i^{\max}$), there exists no local path that reaches the outside of the sensing area of robot $i$. In this case, the TPP (\ref{optim}) is considered infeasible, robot $i$ switches to \textbf{Emerg} mode and the braking controller (\ref{bra_u}) is applied. When robot $i$ is in \textbf{Emerg} mode, it will continue monitoring the environment (by updating $\mathcal{T}_{c,i}^L$). Once a feasible local path is found, it will switch back to \textbf{Free} mode. The whole motion coordination process is summarized in Algorithm 6.

\begin{algorithm}\label{algorithm6}
\caption{\textit{motionCoordination}}
\begin{algorithmic}[1]
\Require (offline) $M_i(\hat X_l), \forall \hat X_l\in \Xi$, $\mathcal{T}_{c, i}$, $\mathsf{B}_i, \mathcal{P}_i$, and $V_{\mathcal{P}_{i}}$.
\Ensure Real-time plan ${\bm \xi}_i^+([t_k, \infty)), t_k\ge 0$.
\State Initialize: ${\bm \xi}_i^-([0, \infty))\leftarrow{\bm \xi}_i^0$, $B_i(\xi_i(0))\leftarrow \beta_{\mathcal{P}_i}(\xi_i(0))$, $ \texttt{Post}(B_i(\xi_i(0)))\leftarrow \cup_{s_i\in B_i(\xi_i(0))} \texttt{Post}(s_i),$ and Robot $i$ is in \textbf{Free} mode.
\While {$t_k> 0$ and $\varphi_i$ is not completed},
\State \parbox[t]{\dimexpr\linewidth-\algorithmicindent}{Compute $B_i(\xi_i(t_k))$ and $\texttt{Post}(B_i(\xi_i(t_{k})))$ according to (\ref{Bi}) and (\ref{PostBi}),}
\State $\tilde{\mathcal{N}}_i(t_k)\leftarrow \textit{conflictDetection}()$,
\If {$\tilde{\mathcal{N}}_i(t_k)\neq \emptyset$}
\State Robot $i$ switches to \textbf{Busy} mode,
\State $\mathcal{Y}_i(t)\leftarrow \textit{planningOrderAssignment}()$,
\State $(\mathcal{T}_{c,i}^L, \xi_i^f)\leftarrow \textit{localTrajectoryGeneration}()$,
\If {$\xi_i^f\neq \emptyset$,}
\State $\bm{\rho}_i\leftarrow \textit{globalTrajectoryGeneration}()$,
\State \parbox[t]{\dimexpr\linewidth-\algorithmicindent}{${\bm \xi}_i^+([t_k, \infty))\leftarrow \textit{DijksTargets}(\mathcal{T}_{c,i}^L, \xi_i(t_k), \xi_i^f)\uplus \bm{\rho}_i$,}
\State Robot $i$ switches to \textbf{Free} mode,
\Else
\State Robot $i$ switches to \textbf{Emerg} mode,
\State ${\bm u}_i\leftarrow {\bm u}_i^{\text{br}}(\xi_i(t_k))$,
\State ${\bm \xi}_i^+([t_k, \infty))\leftarrow {\bm \xi}_i(\xi_i(t_k), {\bm u}_i, [0, \infty))$,
\EndIf
\Else
\State ${\bm \xi}_i^+([t_k, \infty))\leftarrow {\bm \xi}_i^-([t_k, \infty))$,
\EndIf
\EndWhile
\end{algorithmic}
\end{algorithm}

The following result shows hat safety is guaranteed under all circumstances.

\begin{theorem}[\bf{Safety}]
  If the sensing radius of the robots satisfies $R>2\max_{i\in \mathcal{V}}\{D_i^{\text{br}}+\Delta \max_{p_i\in \mathbb{W}}\{\|\dot p_i\|\}\}$, where $\Delta$ is the conflict detection interval, then the resulting real-time plan of Algorithm 6 guarantees that there will be no obstacle-robot and inter-robot collisions for robot $i$.
\end{theorem}

\textbf{\emph{Proof:}} If the TPP (\ref{optim}) is feasible (robot $i$ will be in \textbf{Free} mode after the trajectory planning process), the local trajectory generation algorithm (Algorithm 3) and the global trajectory generation algorithm (Algorithm 5) guarantees that the constraints (\ref{optim-a}) and (\ref{optim-b}) of (\ref{optim}) are satisfied. That is to say, there will be no obstacle-robot and inter-robot collisions for robot $i$. If the TPP (\ref{optim}) is infeasible (robot $i$ will switch to \textbf{Emerg} mode), robot $i$ would apply the braking controller ${\bm u}_i^\text{br}$ until it stops. Since $R>2\max_{i\in \mathcal{V}}\{D_i^{\text{br}}+\Delta \max_{p_i\in \mathbb{W}}\{\|\dot p_i\|\}\}$, it guarantees that conflict between any pair of robots $(i, j)$ will be detected at the time that both robot $i$ and $j$ are outside of the braking area of the other. This means that there will be no inter-robot collision during the emergency stop process. On the other hand, when constructing $\mathcal{T}_{i}$, one has $\psi_i(\texttt{proj}_l(p_i))\subseteq \mathbb{F}, \forall p_i\in \hat{\mathbb{X}}_i$), \textit{i.e.,} $\texttt{dist}(p_i, \mathbb{O})\ge D_i^{\text{br}}, \forall p_i\in \hat{\mathbb{X}}_i$. Moreover, in the replanning process, it also requires that the distance between $\texttt{proj}_l([\xi_n, \xi_r])$ and static obstacles $\mathbb{O}$ is no less than  $D_i^{\text{br}}$ (line 10, Algorithm 3). Therefore, there will be no obstacle-robot collision during the emergency stop process. Thus, no collision will occur for the whole process.
$\square$

\begin{theorem}[\bf{Correctness}]
  Let $\bm{\xi}_i^{\text{rt}}([0, \infty))$ be the real-time moving trajectory of robot $i$. Then, $\bm{\xi}_i^{\text{rt}}([0, \infty))\models \varphi_i$ if robot $i$ never enters the \textbf{Emerg} mode or stays in \textbf{Emerg} mode for a finite time.
\end{theorem}

\textbf{\emph{Proof:}} If robot $i$ never enters the \textbf{Emerg} mode, then one has from Algorithm 6 that robot $i$ never detects a conflict or the local trajectory generation algorithm (Algorithm 3) is always feasible for robot $i$. Then, one can conclude that $\bm{\xi}_i^{\text{rt}}([0, \infty))\models \varphi_i$. Otherwise, since robot $i$ stays in \textbf{Emerg} mode for a finite time, it means that robot $i$ can always switch back to \textbf{Free} mode, which implies the existence of a satisfying trajectory. In the light of Algorithm 6, the real-time moving trajectory $\bm{\xi}_i^{\text{rt}}([0, \infty))\models \varphi_i$. $\square$
%

\begin{remark}
Due to the distributed fashion of the solution and the locally available information, the proposed motion coordination strategy is totally scalable in the sense that the computational complexity of the solution is not increasing with the number of robots. In addition, it is straightforward to extend the work to MRS scenarios where unknown static obstacles and moving obstacles are presented.
\end{remark}


\begin{remark}\label{remark:completeness}
The workspace discretization, the planning order assignment, and the over-approximation of the braking area and time influence the completeness (the ability to find a solution when one exists) of the proposed solution. To improve completeness, online refinement of the discretization of the workspace, searching the space of the prioritization scheme (\textit{e.g.,} the randomized search approach with hill-climbing \cite{bennewitz2002finding}), and less-conservative approximation of the braking area and time (using the real-time non-position information) can be utilized. The disadvantage is that the resulting approach will be computationally more complex.
\end{remark}

\subsection{Discussion of livelocks and deadlocks}

In this subsection, the possibilities for livelocks and deadlocks are discussed. Firstly, the definitions of deadlock and livelock are given.

\begin{definition}
We say that a robot $i$ is in a \emph{deadlock} if it remains in \textbf{Emerg} mode for an indefinite period (stop forever). We say that a robot $i$ is in a \emph{livelock} if it is in \textbf{Free} mode (thus can move in the workspace) but unable to satisfy its specification.
\end{definition}

Livelocks never occur. The reason is that according to our motion coordination algorithm (Algorithm 6), if a robot $i$ finds during execution that its specification is not satisfiable, then it will switch to \textbf{Emerg} mode and thus comes to a stop. In addition, it switches back to \textbf{Free} mode only when the task becomes satisfiable again. Therefore, there will be no livelocks.

Before discussing deadlocks, the definition of mutual satisfiability is provided. Denote by $\mathcal{P}_G:=\mathcal{P}_1\otimes\ldots\otimes \mathcal{P}_{|\mathcal{V}|}$ the global PBA, which is the product of all local PBAs $\mathcal{P}_i, i\in \mathcal{V}$.

\begin{definition}
  We say that the set of LTL specifications $\{\varphi_1, \ldots, \varphi_{|\mathcal{V}|}\}$ is \emph{mutually satisfiable} if there exists an accepting run $\bm{p}_G$ of $\mathcal{P}_G$ such that
  \begin{eqnarray*}
    &&\phi_i(\texttt{proj}_l({\xi}_i(t)))\cap \phi_j(\texttt{proj}_l(\xi_j(t))) = \emptyset,\\ &&\hspace{3.5cm}\forall i, j\in \mathcal{V}, i\neq j, \forall t\ge 0,
  \end{eqnarray*}
where $\bm{\xi}_i$ is the projection of $\bm{p}_G$ onto the local CTS $\mathcal{T}_i$.
\end{definition}

When the set of LTL specifications $\{\varphi_1, \ldots, \varphi_{|\mathcal{V}|}\}$ is not mutually satisfiable, deadlocks can happen because the LTL specifications of two or multiple robots can not be satisfied simultaneously in a collision-free manner. This type of deadlocks is unresolvable. When the set of LTL specifications $\{\varphi_1, \ldots, \varphi_{|\mathcal{V}|}\}$ is mutually satisfiable, deadlocks can also happen due to the incompleteness of the motion coordination algorithm (Remark \ref{remark:completeness}) or the locally available information of the neighboring robots. Existing methods such as path refinement \cite{atia2018ogpr} and simultaneous trajectory planning \cite{molloy2019simultaneous} (instead of the sequential trajectory planning used in this work), can be invoked to resolve this type of deadlocks. To detect a possible deadlock, one can a priori specify a maximum time $t_{\max}$ allowed for a robot to stay in \textbf{Emerg} mode.


\section{Case studies}

In this section, two case studies are provided  to illustrate the effectiveness and computational tractability of the proposed framework. All simulations are carried out in Matlab 2018b on a Dell laptop with Windows 10, Intel i7-6600U CPU2.80GHz and 16.0 GB RAM.

\subsection{Example 1}

Consider a MRS consisting of $N=4$ robots, the dynamics of robot $i$ is given by (\ref{ex:dynamics}), and the velocity and input constraints are given by (\ref{ex:cons}). For robots $i\in \{1, 2\}$, one has that $v_{i, \max}= 1 {\rm  m/s}$, $\omega_{i, \max}=0.5 {\rm rad/s}$, and $a_{i, \max}= 2  {\rm m/s^2}$. For robots $i\in \{3, 4\}$, one has that $v_{i, \max}= 1 {\rm  m/s}$, $\omega_{i, \max}=0.5 {\rm rad/s}$, and $a_{i, \max}= 1.5 {\rm m/s^2}$. Then, using the braking controller (\ref{brakingcontroller1}), one can get that $T_i^{\text{br}}=0.5 {\rm s}, D_i^{\text{br}}=0.25 {\rm m}, i=1, 2$ and $T_i^{\text{br}}=0.5 {\rm s}, D_i^{\text{br}}=0.33 {\rm m}, i=3, 4$. The sensing radius of each robot is $R=3.5 {\rm m}$ and the conflict detection period is $\Delta=0.1 {\rm s}$. Then one has that $R>2\max_{i\in \mathcal{V}}\{D_i^{\text{br}}+\Delta v_{i, \max}\}$.

\begin{figure}[h!]
\centering
\includegraphics[width=.4\textwidth]{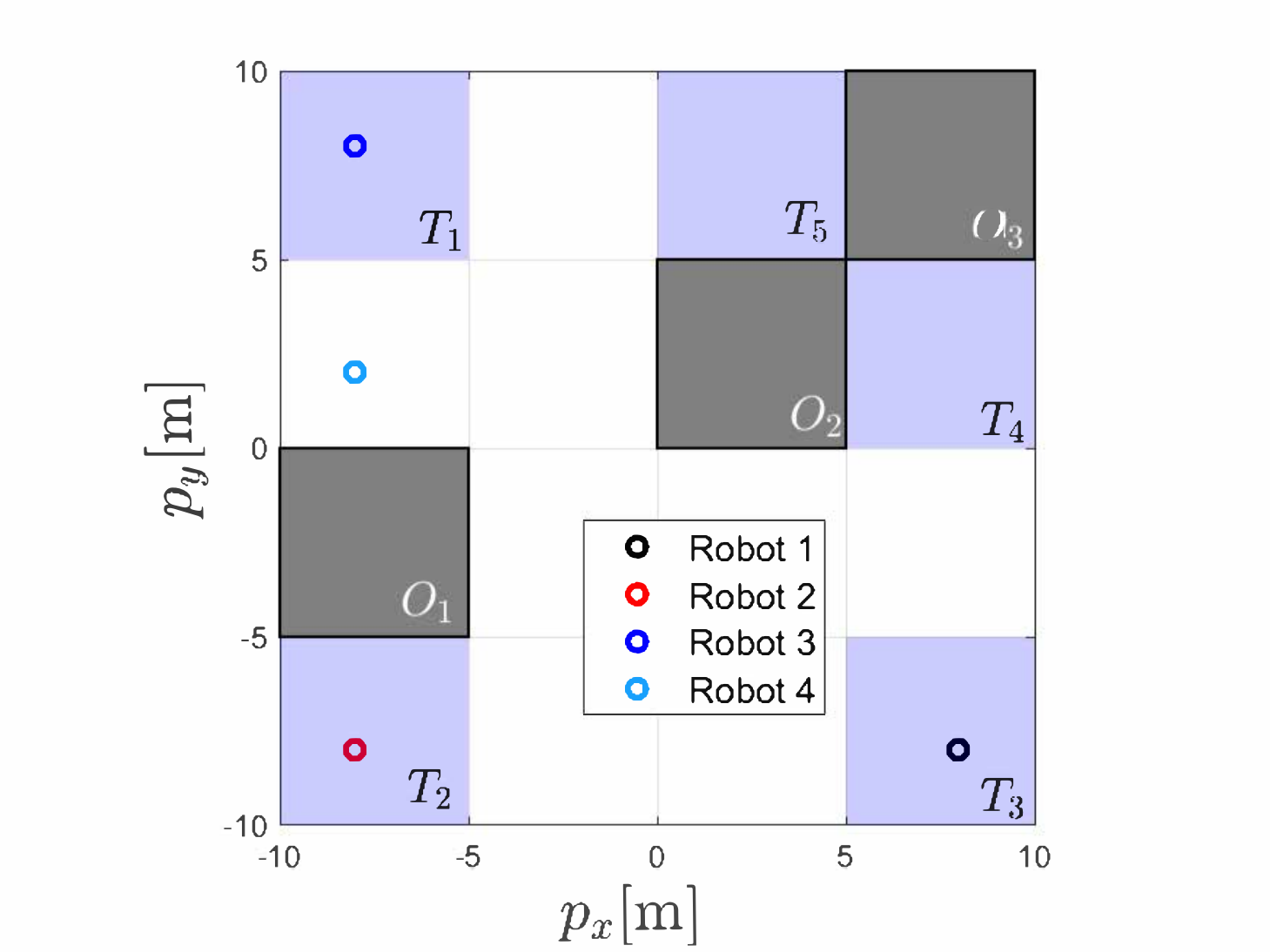}
\caption{The workspace for the group of robots in Example 1. }\label{fig4}
\end{figure}

As shown in Fig. \ref{fig4}, the workspace $\mathbb{W}$ we consider is a $20\times 20 {\rm m^2}$ square, where the gray areas, marked as $O_1, O_2, O_3$, represent three static obstacles, the colored small circles represent the initial position of the robots, and the light blue areas, marked as $T_1, T_2, \ldots, T_5$, represent a set of target regions in the workspace. Initially, $\theta_i=0, v_i(0)=0, \forall i$.

Each robot is assigned to persistently survey two of the target regions in the workspace. In LTL formulas, the specification for each robot is given by
\begin{itemize}
  \item $\varphi_1=\square (\mathbb{W}\wedge \neg \mathbb{O})\wedge\square\lozenge T_1\wedge\square\lozenge T_2$,
  \item $\varphi_2=\square (\mathbb{W}\wedge \neg \mathbb{O})\wedge\square\lozenge T_1\wedge\square\lozenge T_5$,
  \item $\varphi_3=\square (\mathbb{W}\wedge \neg \mathbb{O})\wedge\square\lozenge T_2\wedge\square\lozenge T_4$,
  \item $\varphi_4=\square (\mathbb{W}\wedge \neg \mathbb{O})\wedge\square\lozenge T_3\wedge\square\lozenge T_5$,
\end{itemize}
where $\mathbb{O}=O_1\cup O_2\cup O_3$. Firstly, a rough cell decomposition is given over the workspace $\mathbb{W}$ such that Assumption \ref{ass3} is satisfied. Then, for each cell $X_l\subset \mathbb{W}\setminus \mathbb{O}$, a grid representation with grid size $0.5 {\rm m}$ is implemented as a finer workspace discretization. The NBA $\mathsf{B}_i$ associated with $\varphi_i, \forall i$ has 3 states and 8 edges by \cite{gastin2001}. The CTS $\mathcal{T}_i$ and the PBA $\mathcal{P}_i$ for each robot are constructed using LTLCon toolbox \cite{kloetzer2005ltlcon}. Finally, the map $M_i$ (defined in (\ref{map})) and potential function for each $\mathcal{P}_i$ (Definition \ref{potentialfunctionforproductautomata}) are computed.

\begin{figure*}[t]
\centering
\subfigure[]{
\includegraphics[width=.45\linewidth]{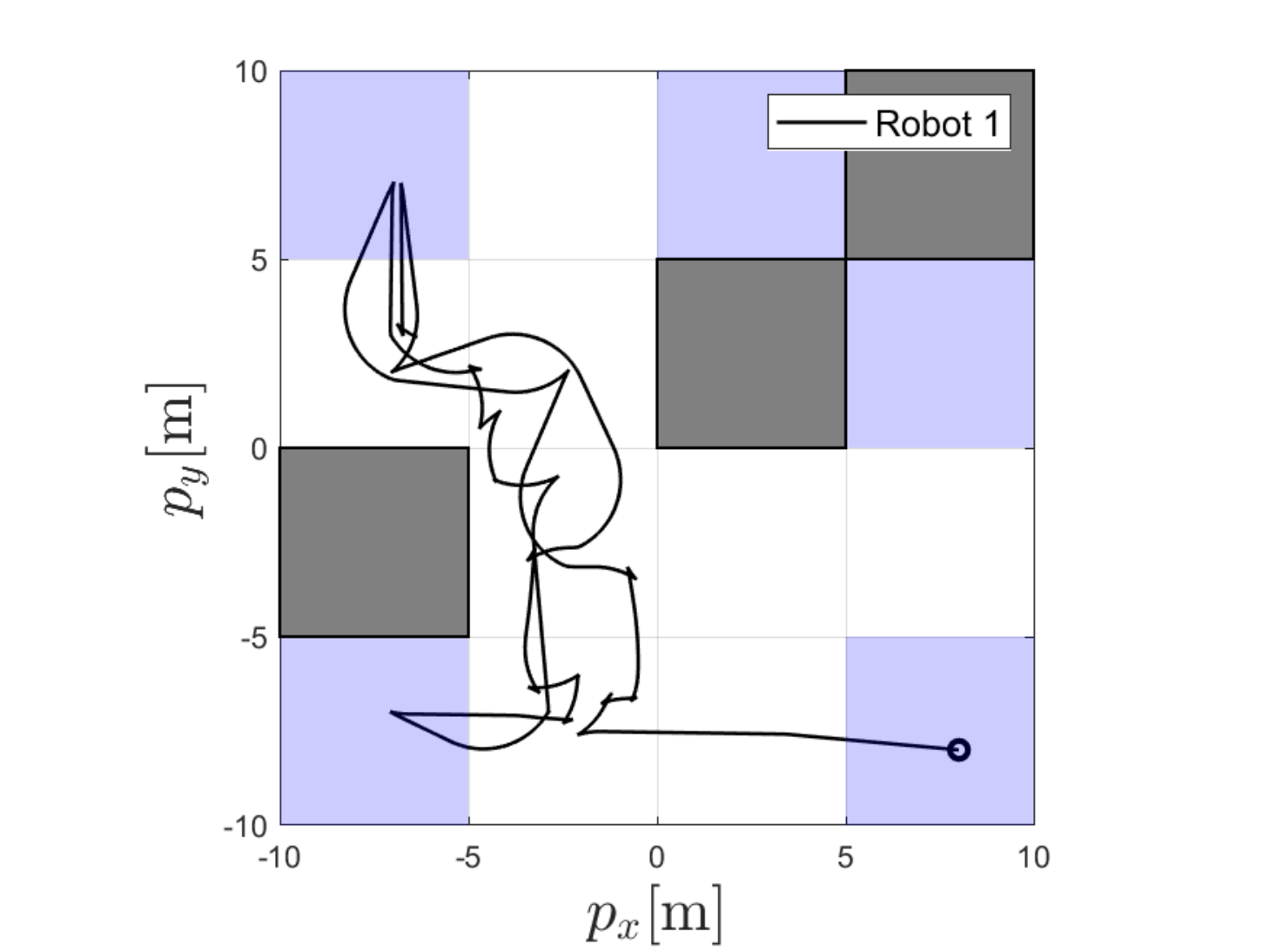}\label{fig5a}
}
\subfigure[]{
\includegraphics[width=.45\linewidth]{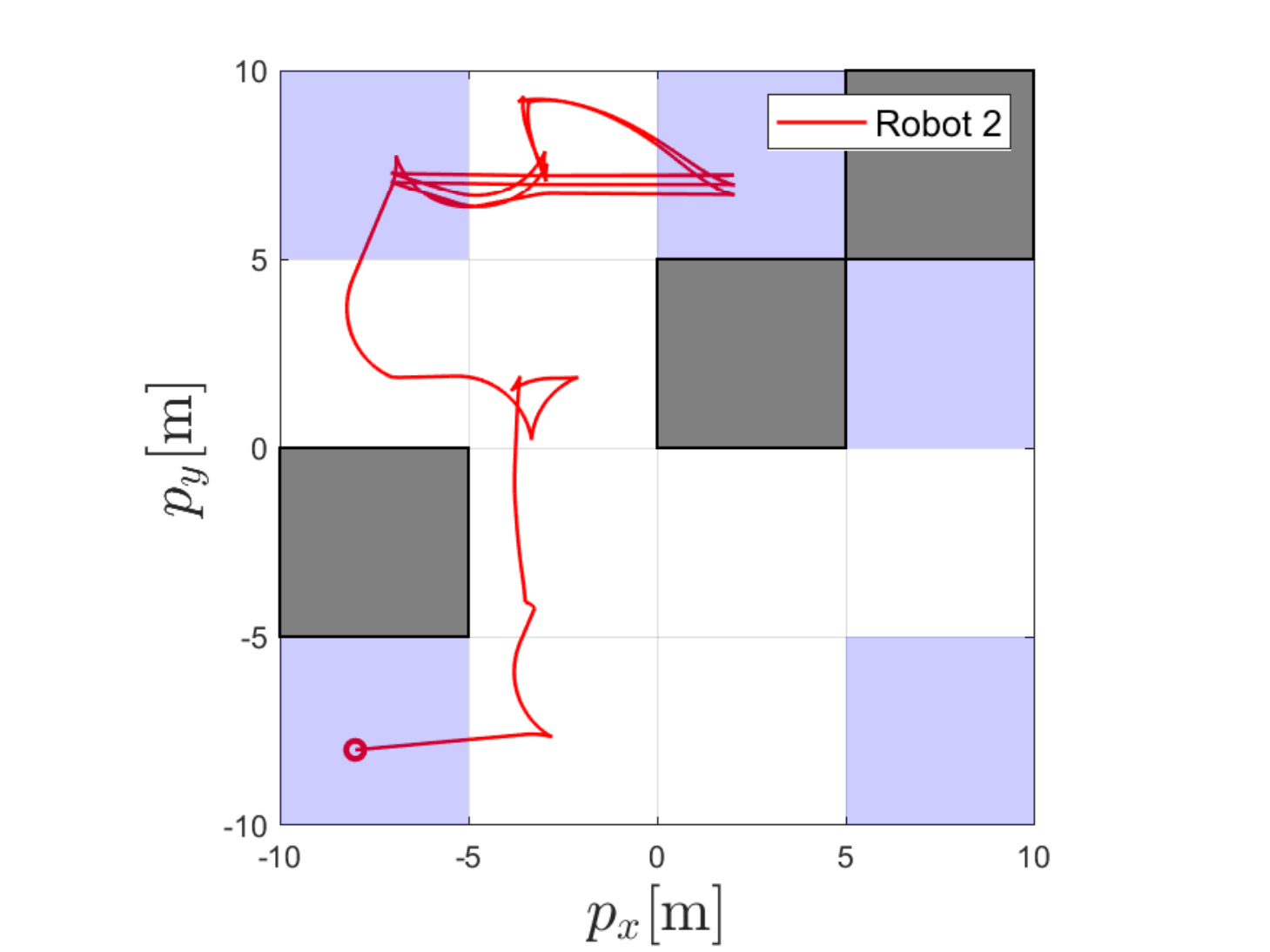}\label{fig5b}
}
\subfigure[]{
\includegraphics[width=.45\linewidth]{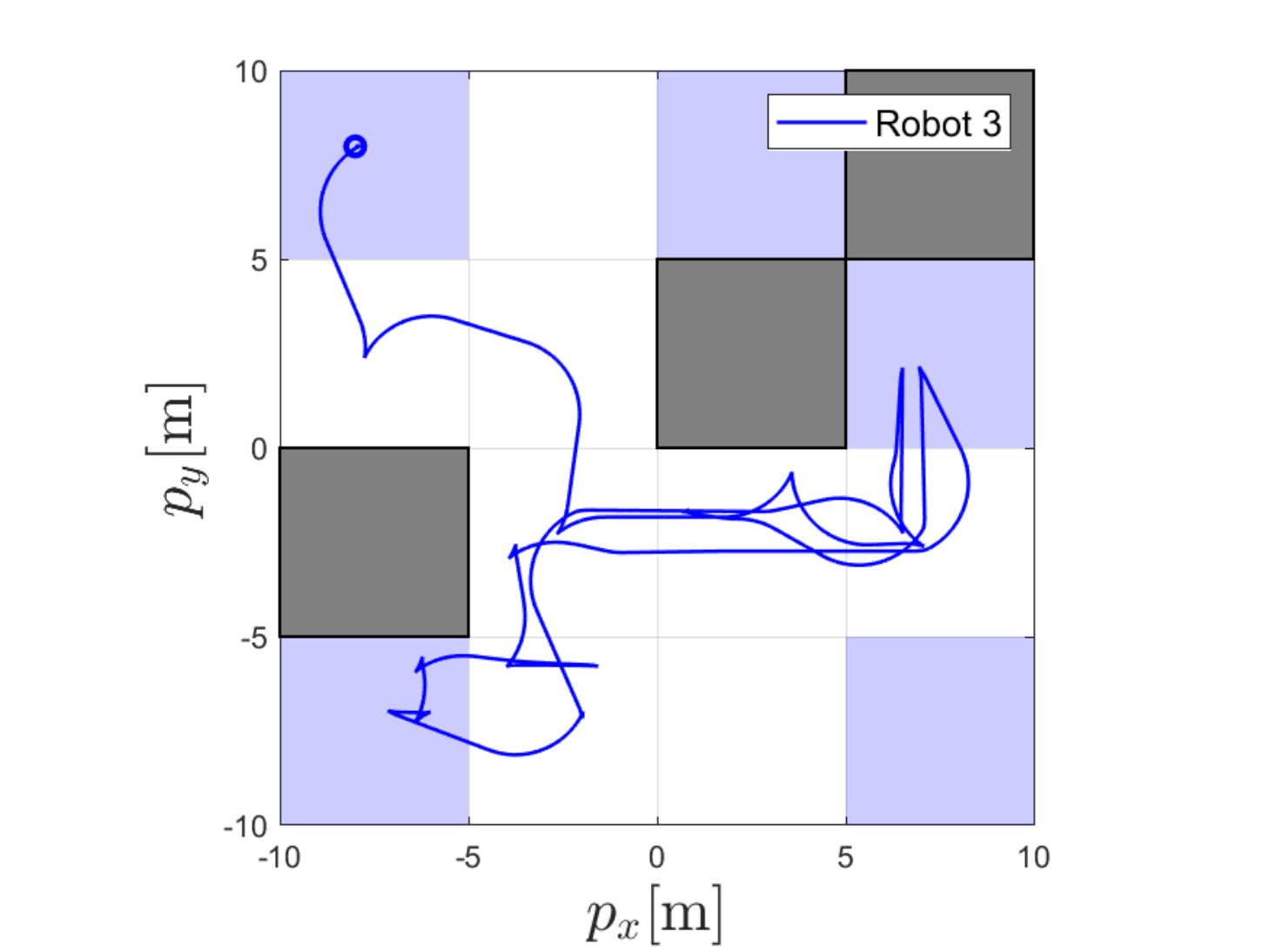}\label{fig5a}
}
\subfigure[]{
\includegraphics[width=.45\linewidth]{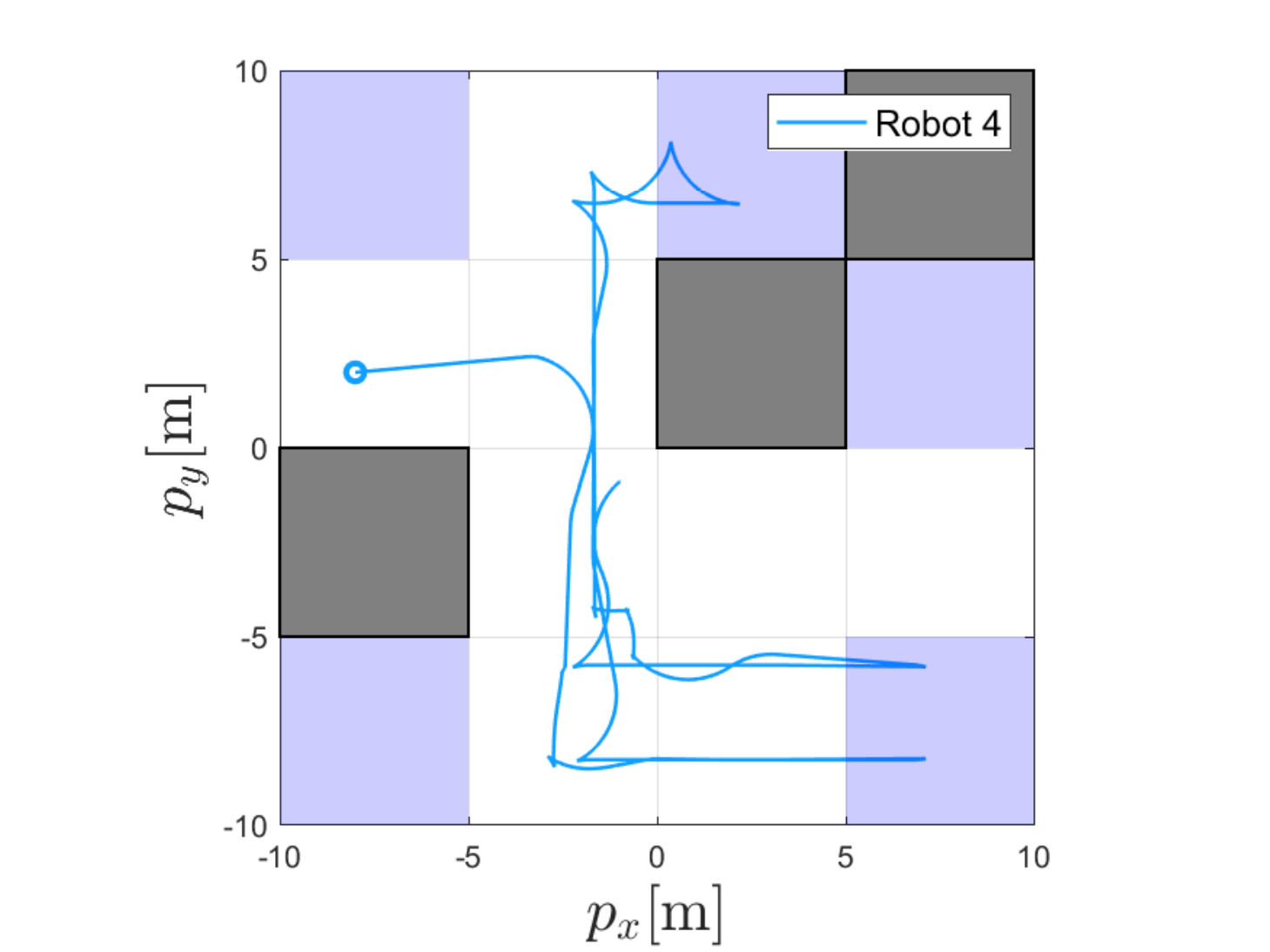}\label{fig5b}
}
\caption{The real-time position trajectory of each robot, where the small circle represents the initial position of each robot.}
\label{fig5}
\end{figure*}

\begin{figure}[H]
\centering
\includegraphics[width=.5\textwidth]{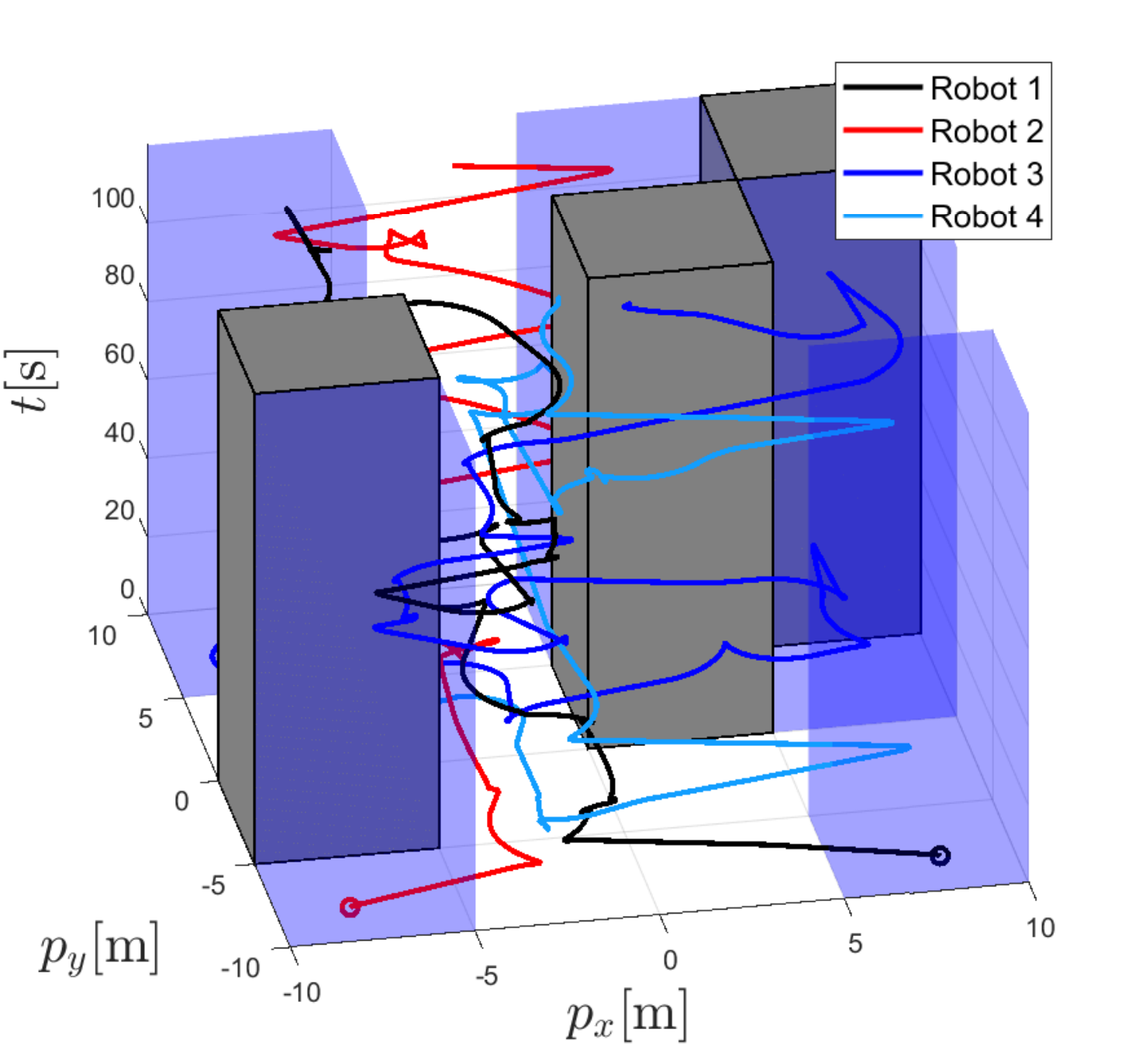}
\caption{The evolution of the position
trajectories with respect to time.}\label{fig6}
\end{figure}

\begin{figure*}[t]
\centering
\subfigure[]{
\includegraphics[width=.45\linewidth]{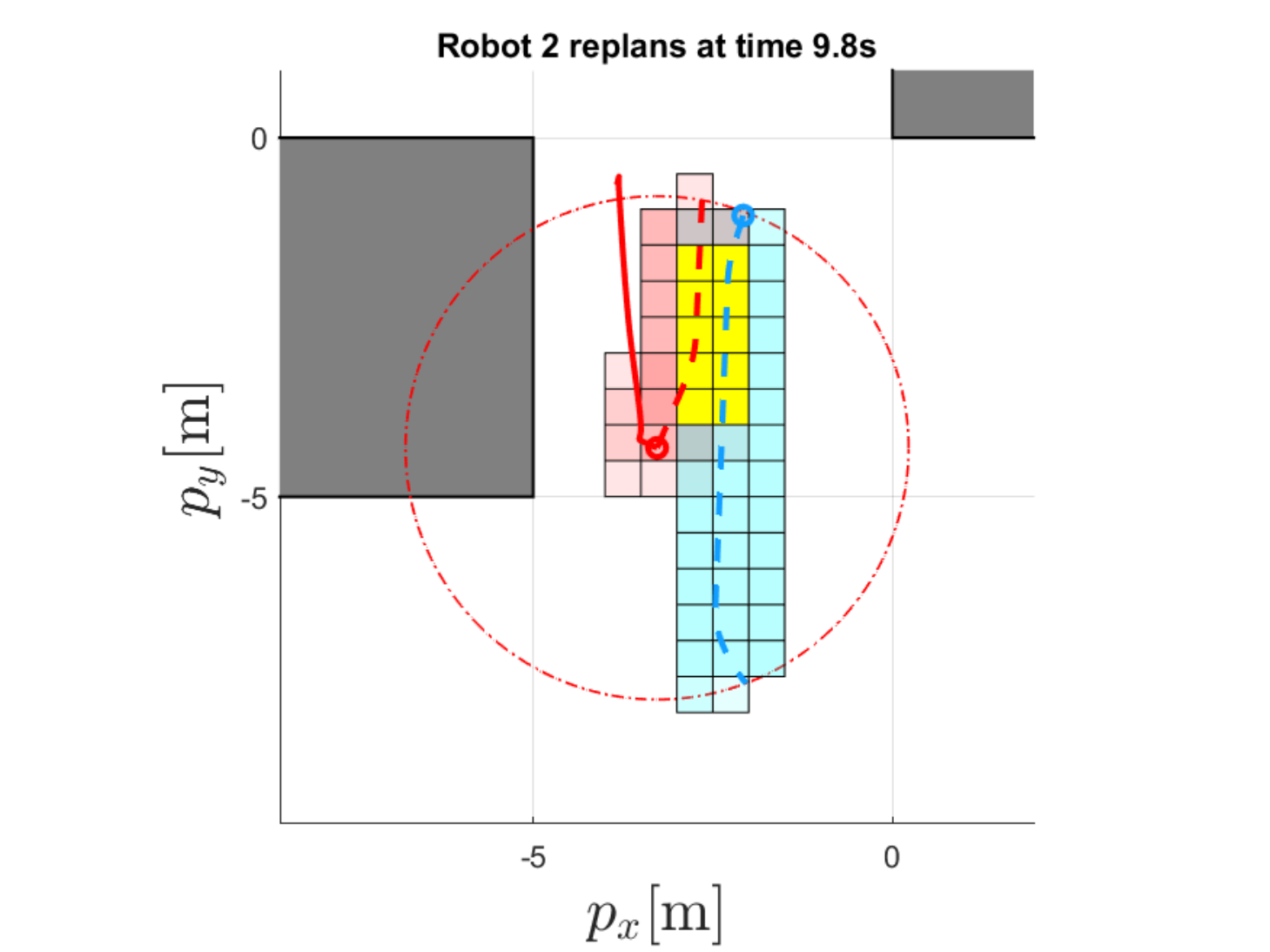}\label{fig5a}
}
\hspace{0.4cm}
\subfigure[]{
\includegraphics[width=.45\linewidth]{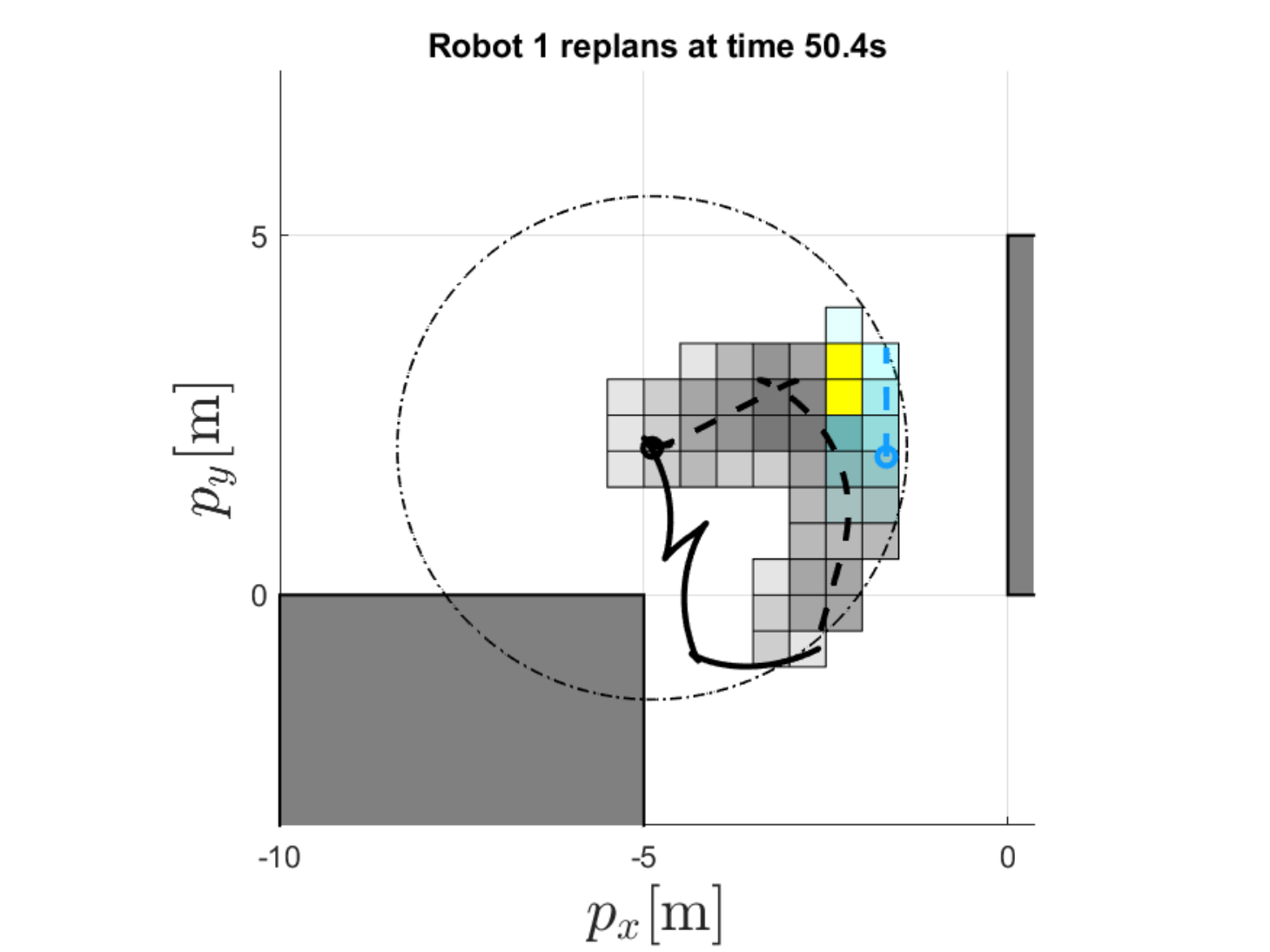}\label{fig5b}
}
\subfigure[]{
\includegraphics[width=.45\linewidth]{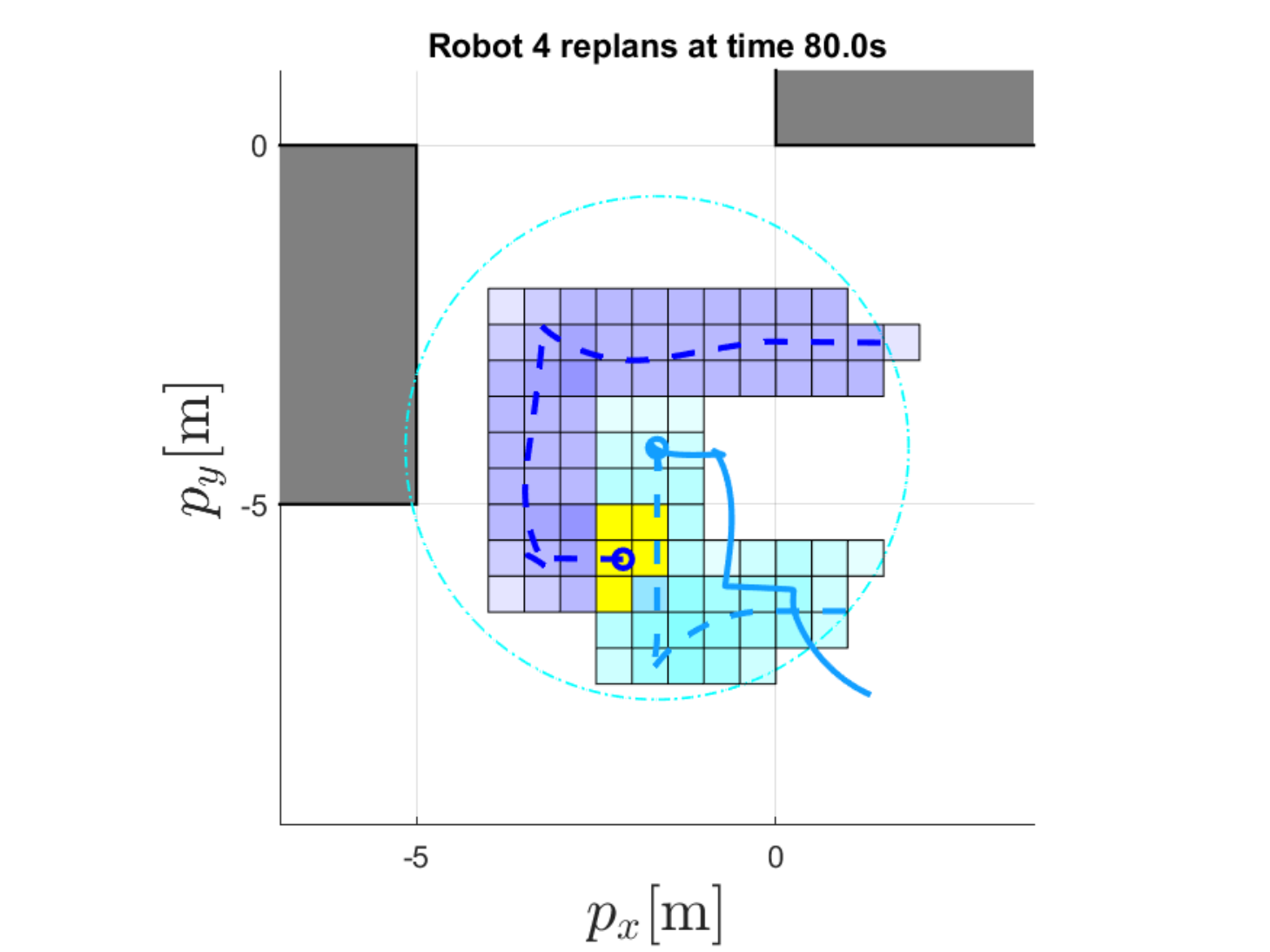}\label{fig5a}
}
\hspace{0.4cm}
\subfigure[]{
\includegraphics[width=.45\linewidth]{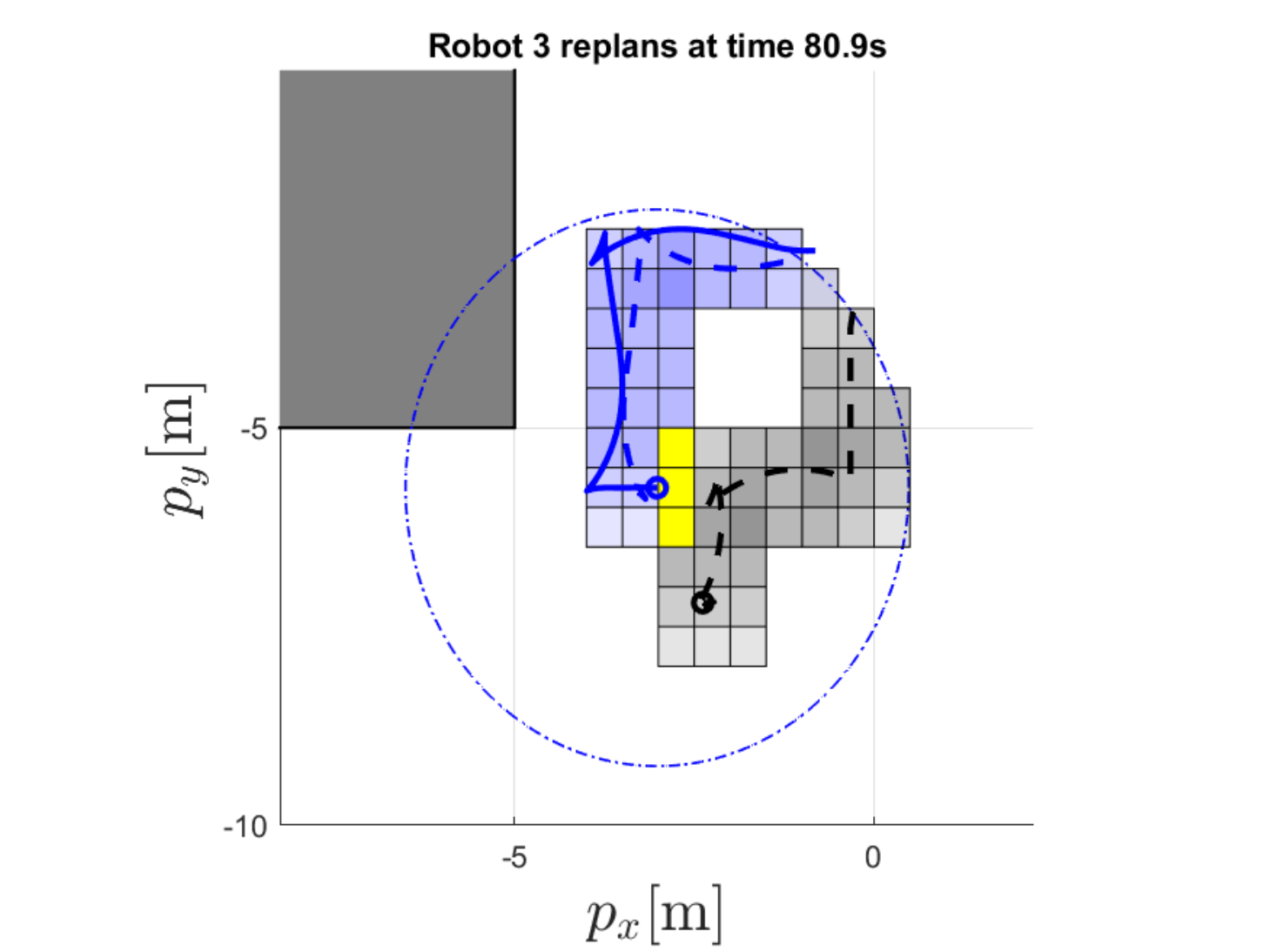}\label{fig5a}
}
\caption{The local motion replanning between different conflict robots. The circle, marked with a dotted line, represents the sensing area of the robot that replans, the dashed lines represent the planned local position trajectories of the robots, the colored small squares represent its (over-approximated) braking areas, the yellow squares represent the conflict region, and the solid line represents the replanned local position trajectory.}
\label{fig7}
\end{figure*}

Initially, each robot $i$ finds a safely satisfying trajectory $\bm{\xi}_i^0$ (in the form of prefix-suffix). Then, the motion coordination algorithm (Algorithm 6) is implemented for each robot during online execution. The real-time position trajectory of each robot is depicted in Fig. \ref{fig5} (a)-(d), respectively. It can be seen that during the simulation time $[0, 120] {\rm s}$, the surveillance task of each robot is satisfied at least once. Fig. \ref{fig6} shows the evolution of the position
trajectories of all robots with respect to time. One can see that the real-time position trajectories are collision-free. During the simulation time, conflicts are detected in total 11 times. Whenever conflicts are detected, the planning order assignment module and the trajectory planning module are activated to resolve the conflicts. In Fig. \ref{fig7} (a)-(d), 4 cases of local motion replanning are depicted, respectively. In each of the subfigures, the circle, marked as dotted line, represents the sensing area of the robot that replans, the dashed lines represent the planned local position trajectories of the robots, the colored small squares represent its (over-approximated) braking areas, the yellow squares represent the conflict region, and the solid line represents the replanned local position trajectory.
A video recording of the real-time implementation can be found here: \url{https://youtu.be/7Xjt62psZe0}.

\begin{figure}[h!]
\centering
\includegraphics[width=.5\textwidth]{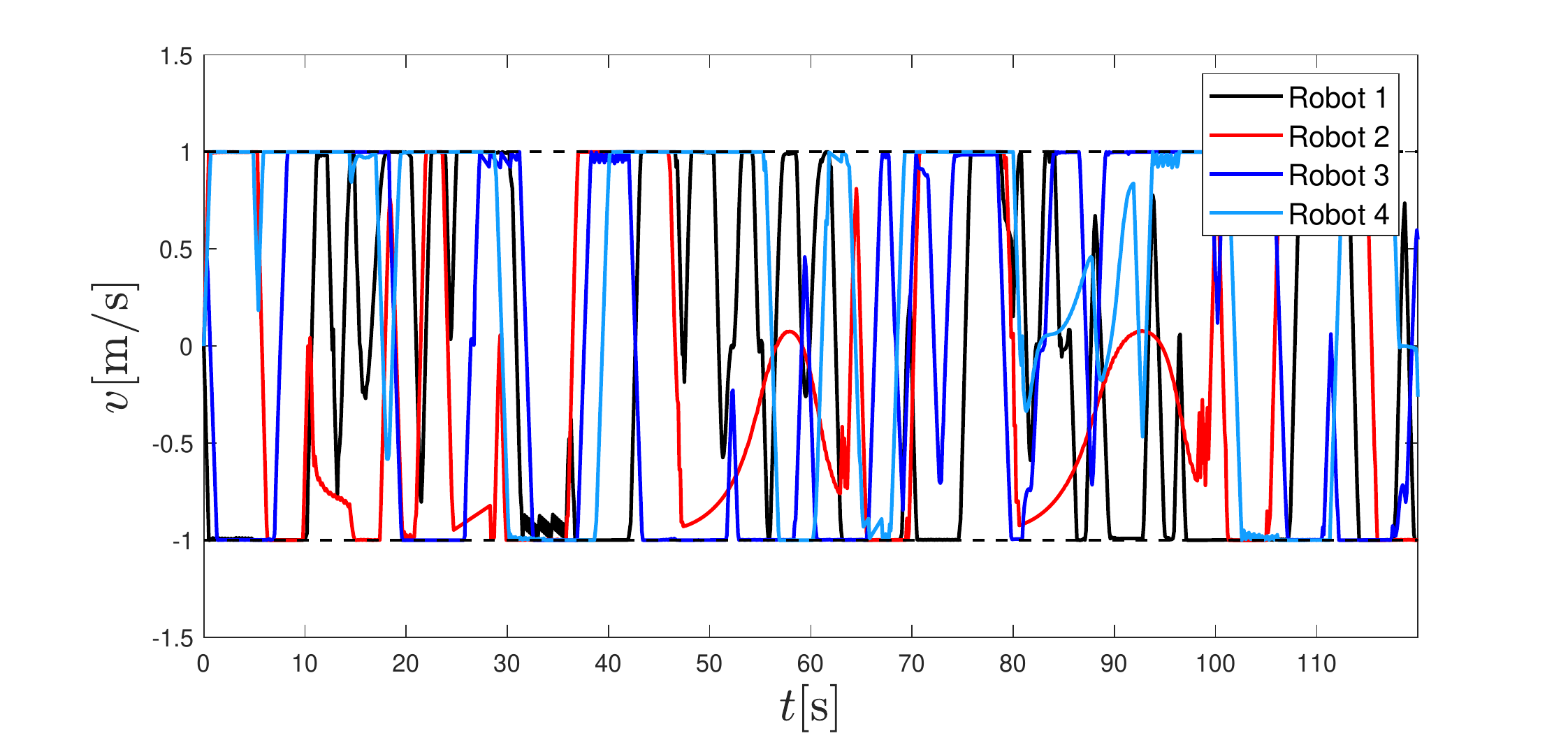}
\caption{The real-time evolution of velocity $v_i$ for each robot, where $v_{i, \max}=1 {\rm m/s}, i\in \{1, 2, 3, 4\}$. }\label{fig8}
\end{figure}

\begin{figure}[H]
\centering
\includegraphics[width=.5\textwidth]{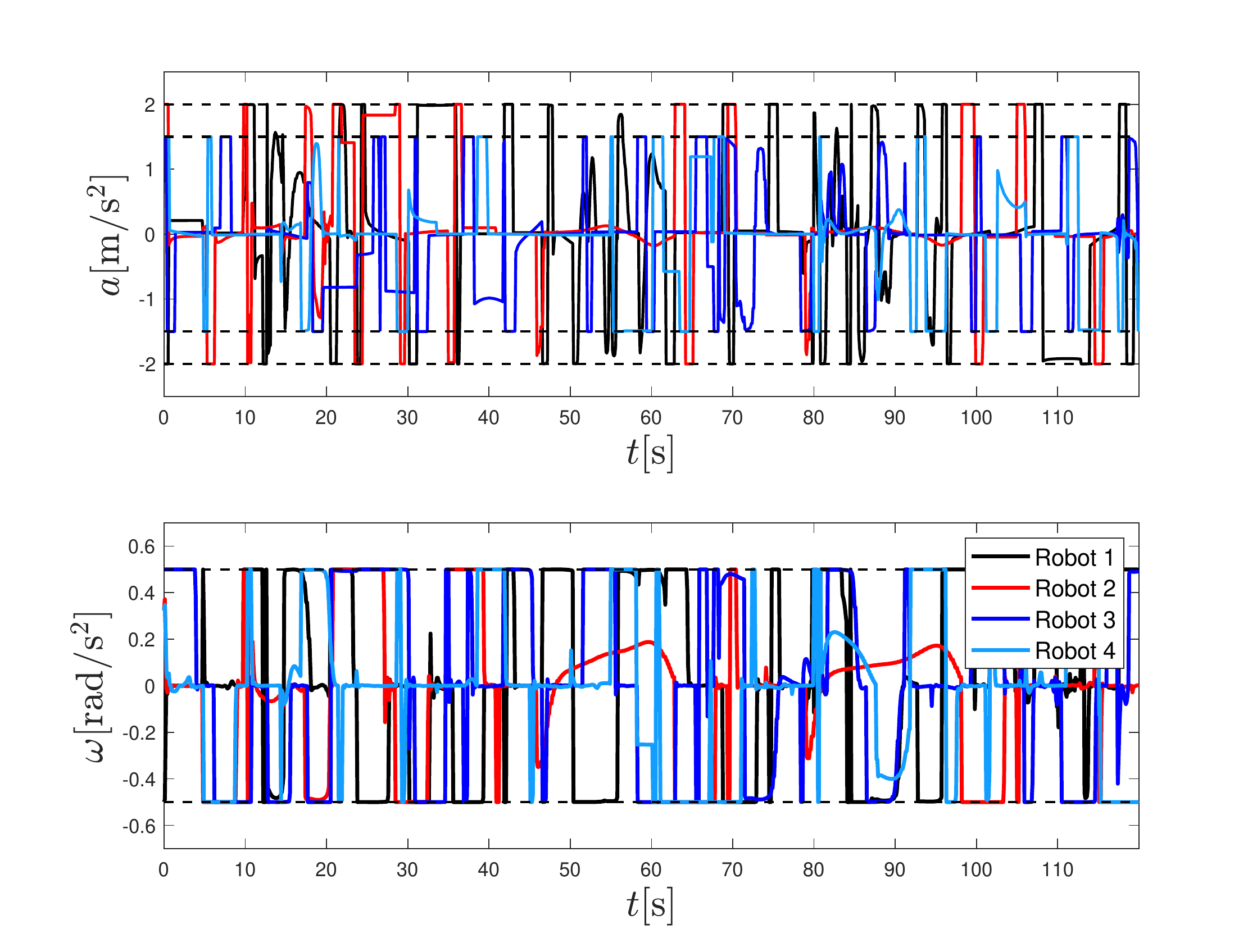}
\caption{The real-time evolution of inputs $(a_i, \omega_i)$ for each robot, where $a_{i, \max} =2  {\rm m/s^2}, \omega_{i, \max} =0.5  {\rm rad/s}, i\in \{1, 2\}$ and $a_{i, \max} =1.5  {\rm m/s^2}, \omega_{i, \max} =0.5  {\rm rad/s}, i\in \{3, 4\}$. } \label{fig9}
\end{figure}

The real-time evolution of velocity $v_i$ and inputs $(a_i, \omega_i)$ of each robot $i$ are plotted in Fig. \ref{fig8} and Fig. \ref{fig9}, respectively. One can see that the velocity constraints and the input constraints are satisfied by all robots at any time.

Finally, we report the computation times of this example. For the 11 local replannings, the average computation time is $1.51 {\rm s}$ and the maximum computation time is $3.30 {\rm s}$. The computation of the global trajectory (Algorithm 5) is quite fast (less than $0.1 {\rm s}$). Note that after the local and global trajectories are obtained, the toolbox ICLOCS2 \cite{nie2018iclocs2} is further employed for trajectory smoothness. In addition, it is worth to point out that we have tried 10 different sets of initial states and specifications for the group of robots, and we find that the initial states and the specifications have little effect on the average time and maximum time of local replanning. There is no change of order of magnitude in the computation time for all trials.

\subsection{Example 2}

This example aims at demonstrating the computational tractability of the approach with respect to the number of robots.

\begin{figure}[h!]
\centering
\includegraphics[width=.4\textwidth]{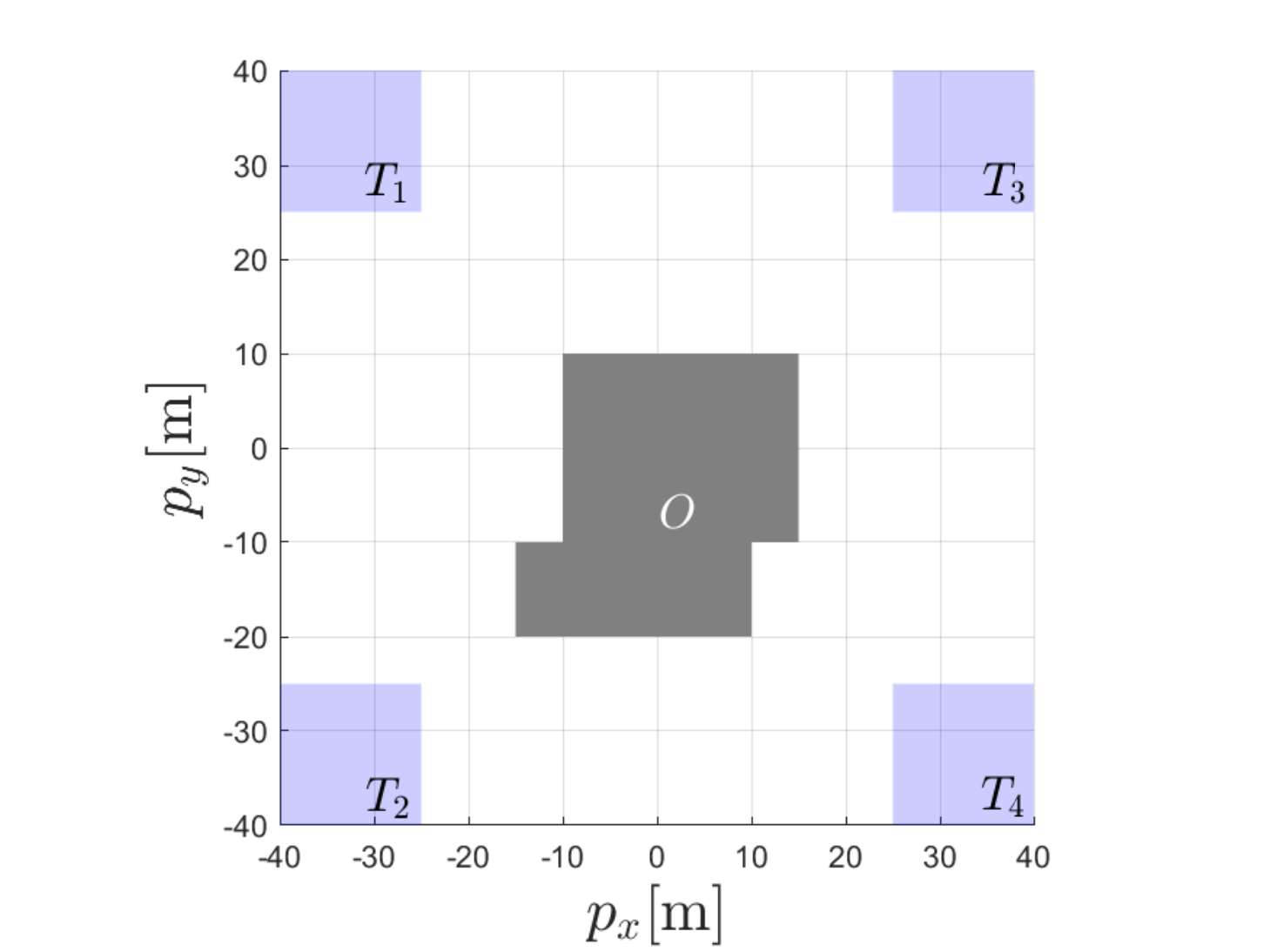}
\caption{The workspace for the group of robots in Example 2. }\label{fig10}
\end{figure}

As depicted in Fig. \ref{fig10}, the workspace $\mathbb{W}$ considered now is a $80\times 80 {\rm m^2}$ square, where the gray area, marked as $O$ is the obstacle, and the light blue areas, marked as $T_1, T_2, \ldots, T_4$, represent 4 target regions in the workspace. We consider respectively the cases where 2, 4, 8 and 16 robots are working in the workspace. The dynamics of robot $i$ is given by a double-integrator, \textit{i.e.,}
\begin{eqnarray*}
  \dot{p}_i &=& v_i \\
  \dot{v}_i &=& u_i, \forall i
\end{eqnarray*}
where $p_i, v_i, u_i\in \mathbb{R}^2$ represent the position, velocity, and input of robot $i$, respectively. The velocity and input of robot $i$ are subject to the hard constraints
\begin{equation*}
  \begin{aligned}
  &||v_{i}(t)||\le v_{i, \max}= 3 {\rm m/s}, \\ &||u_{i}(t)||\le u_{i, \max}= 6 {\rm m/s^2}, \forall i.
  \end{aligned}
\end{equation*}
The braking controller $\bm{u}_i^{\text{br}}$ is designed as
\begin{equation*}
  u_i^\text{br}(t)=\left\{\begin{aligned}
  -u_{i,\max}\frac{v_i(t)}{||v_i(t)||}, &\quad \text{if}\; ||v_i(t)||\neq 0,\\
  0_2 \quad\quad\quad, &\quad \text{if}\; ||v_i(t)||= 0.
  \end{aligned}\right.
\end{equation*}
Then, one can derive the braking time $T_i^{\text{br}}={||v_{i, \max}||}/{||u_{i, \max}||}=0.5  {\rm s}$ and the braking distance $D_i^{\text{br}}=||v_{i, \max}||^2/2||u_{i, \max}||= 0.75  {\rm m}, \forall i$. The sensing radius of each robot is $R=6 {\rm m}$ and the conflict detection period is $\Delta=0.1 {\rm s}$.

For simplicity, we consider that half of the robots are assigned to persistently survey target regions $T_1$ and $T_3$, and the other half are assigned to persistently survey target regions $T_2$ and $T_4$. In LTL formulas, the specification for each robot is given by
\begin{itemize}
  \item $\varphi_i=\square (\mathbb{W}\wedge \neg O)\wedge\square\lozenge T_1\wedge\square\lozenge T_4, i=1,3,\ldots, 15,$
  \item $\varphi_i=\square (\mathbb{W}\wedge \neg O)\wedge\square\lozenge T_2\wedge\square\lozenge T_3, i=2,4,\ldots,16$.
\end{itemize}
A grid representation with grid size $2 {\rm m}$ is implemented as the workspace discretization.

For each of the cases (2, 4, 8 or 16 robots), the simulation time is $150 {\rm s}$. The simulation results are summarized in TABLE II, where the total number of conflict times (CT), the average computation time for local replanning (ATLR), and the maximum computation time for local replanning (MTLR) are reported. It can be seen that as the number of robots grows (exponentially), the CT grows significantly (the reason is that the workspace is the same). The ATLR and MTLR grow, but not significantly. One can see that even when there are 16 robots, the local replanning is still efficient.

\begin{table}[h]
\centering
\caption{The CT, ATLR, and MTLR with respect to the number of robots.}
\begin{tabular}{c||c|c|c}
\hline
number of robots & CT & ATLR ({\rm s}) & MTLR ({\rm s}) \\ \hline
2                & 0  &  -    &  -    \\ \hline
4                & 6  & 0.23 & 2.99 \\ \hline
8                & 19 & 0.31 & 3.80 \\ \hline
16               & 72 & 0.63 & 4.94 \\ \hline
\end{tabular}
\end{table}

\section{Conclusion}

In this paper, the online MRMC problem for a group of mobile robots moving in a shared workspace was considered. Under the assumptions that each robot has only local view and local information, and subject to both state and input constraints, a fully distributed motion coordination strategy was proposed for steering individual robots in a common workspace, where each robot is assigned a LTL specification. It was shown that the proposed strategy can guarantee collision-free motion of each robot. In the future, MRMC under both team and individual tasks will be of interest. In addition, the experimental validation of the proposed strategy by real-world robots will be pursued.

\end{document}